\documentclass{article} 
\usepackage[final]{colm2026_conference}

\usepackage{microtype}
\usepackage{hyperref}
\usepackage{url}
\usepackage{booktabs}
\usepackage{wrapfig}
\usepackage{amsmath}
\usepackage{amssymb}
\usepackage{mathtools}
\usepackage{amsthm}
\usepackage{makecell}
\usepackage{graphicx}
\usepackage{subcaption}
\usepackage{xcolor}
\usepackage{multirow}
\usepackage{multicol}
\usepackage{pifont}
\usepackage[misc]{ifsym}

\usepackage{tcolorbox}
\tcbuselibrary{listings,skins,breakable}

\definecolor{promptbg}{RGB}{248,248,248}
\definecolor{promptframe}{RGB}{200,200,200}

\newtcblisting{promptbox}[1][]{
  listing only,
  colback=promptbg,
  colframe=promptframe,
  boxrule=0.4pt,
  arc=2pt,
  left=6pt,right=6pt,top=4pt,bottom=4pt,
  listing options={
    basicstyle=\ttfamily\footnotesize,
    breaklines=true,
    breakatwhitespace=false,
    columns=fullflexible,
    keepspaces=true,
    breakautoindent=false,   
    breakindent=0pt          
  },
  #1
}

\usepackage{color}
\usepackage[table]{xcolor}
\definecolor{mygreen}{RGB}{0,120,90}

\usepackage[capitalize,noabbrev]{cleveref}

\definecolor{darkblue}{rgb}{0, 0, 0.5}
\hypersetup{colorlinks=true, citecolor=mid, linkcolor=primary, urlcolor=mid}

\renewcommand{\cite}[1]{\citep{#1}}

\title{To Mix or To Merge: Toward Multi-Domain Reinforcement Learning for Large Language Models}

\author{
\mbox{
Haoqing Wang$^{1\dagger}$, Xiang Long$^{1\dagger}$, Ziheng Li$^{2,1\dagger}$, Yilong Xu$^1$, Tingguang Li$^{1}$,
}\\
and Yehui Tang$^{1{~\textrm{\Letter}}}$ \\
$^1$ Samsung Research, Beijing, China \quad\quad $^2$ Peking University\\
\texttt{\{haoqing.wang, yehui.tang\}}@samsung.com \\
$^\dagger$~Equal Contribution\quad\quad$^\textrm{\Letter}$~Corresponding Author
}

\begin{document}

\maketitle
\begin{abstract}
Reinforcement Learning with Verifiable Rewards (RLVR) plays a key role in stimulating the explicit reasoning capability of Large Language Models (LLMs). We can achieve expert-level performance in some specific domains via RLVR, such as coding or math. When a general multi-domain expert-level model is required, we need to carefully consider the collaboration of RLVR across different domains. The current state-of-the-art models mainly employ two different training paradigms for multi-domain RLVR: mixed multi-task RLVR and separate RLVR followed by model merging. However, most of the works did not provide a detailed comparison and analysis about these paradigms. To this end, we choose multiple commonly used high-level tasks (e.g., math, coding, science, instruction following, and agent) as our target domains and design extensive qualitative and quantitative experiments using open-source datasets. We find the RLVR across domains exhibits small mutual interferences, and reasoning-intensive domains have mutually synergistic effects. Furthermore, we analyze the internal mechanisms from the perspectives of information constraints, model prediction behavior and self-verification. Our homepage is at \url{https://github.com/Mosi-AI/M2RL}.
\end{abstract}

\section{Introduction}
Large language models (LLMs) \citep{jaech2024openai,guo2025deepseek,yang2025qwen3} have achieved significant success in various natural language processing (NLP) tasks and more challenging reasoning tasks, i.e., mathematics and software engineering. Extensive pre-training on trillion-token scale corpora is indispensable for the acquisition of comprehensive world knowledge and potential reasoning capabilities. Furthermore, the post-training process serves to stimulate explicit reasoning capabilities and align the model's outputs with human-centric stylistic and structural expectations. During the post-training process, Reinforcement Learning with Verifiable Rewards (RLVR) \citep{zhang2025survey} plays a key role and has gained significant attention \citep{wen2025reinforcement,gao2025soft}.

\begin{figure}[t]
    \centering
    \includegraphics[width=0.8\linewidth]{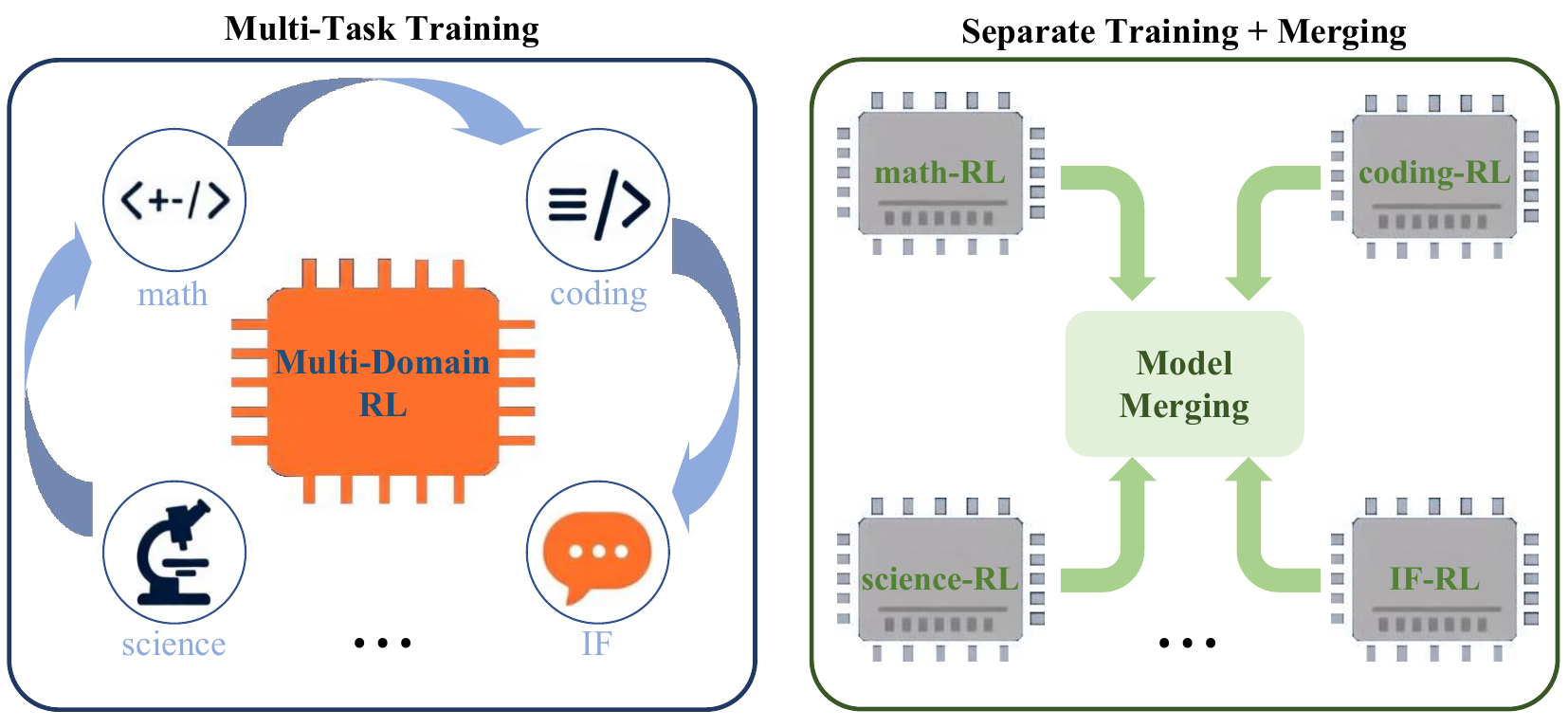}
    \caption{The two training paradigms for multi-domain RLVR: mixed multi-task training and separate training followed by model merging.}
    \label{framework}
\end{figure}

With the help of RLVR, many works have achieved incredibly powerful task solving abilities in some specific domains, such as coding \citep{zhu2024deepseek,hui2024qwen2} and math \citep{yang2024qwen2,shao2024deepseekmath}. When we further want to obtain a general expert-level model that excels at solving tasks from different domains, the cross-domain reinforcement learning is essential. Considering that multi-task reinforcement learning may encounter gradient interference \citep{bai2023picor,wu2025imbalanced}, it is important to deeply analyze the collaboration of multi-domain RLVR. The existing state-of-the-art models \citep{guo2025deepseek,yang2025qwen3,zeng2025glm,xiao2026mimo} typically apply two training paradigms: 1) mixed multi-task RLVR learns based on heterogeneous rewards from different domains simultaneously; 2) separate domain-specific reinforcement learning and then merging different expert models with weight merging \citep{hitit2025systematic} or distillation \citep{agarwal2024policy}. However, most of these works did not share key insights about the comparison between the two training paradigms and their internal mechanisms. In this work, we aim to fill this gap by detailed comparisons and analyses.

We mainly examine five common RLVR domains: math, coding, science, instruction following, and agent. The Qwen3-4B-Base model \citep{yang2025qwen3} is used as the initial model of post-training for both reliability and operability. We use the open-source datasets from Nemotron 3 Nano \citep{blakeman2025nemotron} for both supervised fine-tuning and reinforcement learning. The final training datasets follow the data blend proportion in their technical report. When we obtain multiple domain-specific expert models, we consider both the weight merging methods (i.e., average merging \citep{wortsman2022model}, task arithmetic merging \citep{ilharco2022editing}, Ties-merging \citep{yadav2023ties} and SCE \citep{wan2025fusechat}) and multi-teacher off/on-policy distillation \citep{agarwal2024policy} for merging them. The framework illustrate of these two training paradigms are shown in Figure \ref{framework}.

We compare the obtained models at multiple benchmarks across five domains using Avg@K metric, and analyze the internal mechanisms from the perspectives of information constraints, model prediction behavior and self-verification. The main findings are as follows:
\begin{itemize}
\item The mixed multi-task RLVR can achieve comparable performance with the separate RLVR followed by model merging with $58.3\%$ GPU hours. The cross-domain RLVR manifests little inter-task interference, particularly between the reasoning-intensive domains, where the synergistic effects are observed.
\item We employ KL divergence as a metric to investigate mechanism behind multi-domain merging methods. We observe that the neighborhood policy transfer during weight merging shapes the domain policies toward the optimal policy, thereby enhancing the performance.
\item Weight merging primarily inherits the original capabilities of the single-task models, whereas the capabilities learned through multi-task training and on-policy distillation exhibit a larger divergence from those learned via single-task training.
\item RLVR induces an emergent self-verification capability that is highly sensitive to task characteristics and training paradigms. The agentic, multi-turn interactions serve as a catalyst for robust process-level verification. Although multi-task RLVR is more sufficient, it induces a severe degradation in process verification capability.
\end{itemize}

\section{Related works}
\subsection{Reinforcement Learning with Verifiable Rewards}
The introduction of DeepSeek-R1 \citep{guo2025deepseek} has brought a widespread and rapid expansion of research into the Reinforcement Learning with Verifiable Rewards (RLVR) paradigm \citep{zhang2025survey}. These works have been comprehensive and explored numerous critical aspects of implementation, such as reward design \citep{albalak2025big,chen2025r1,lambert2024tulu}, policy optimization \citep{shao2024deepseekmath,zheng2025group,yu2025dapo}, sampling strategy \citep{cui2025process,dong2025agentic} and various insightful observations \citep{yue2025does}.

Recent studies have utilized RLVR to achieve expert-level performance in some specific domains, such as coding \citep{zhu2024deepseek,hui2024qwen2} and math \citep{yang2024qwen2,shao2024deepseekmath}. However, the fusion of these disparate reinforcement learning domains into a general expert-level model remains an open question. DeepSeek-R1 \citep{guo2025deepseek} and Qwen3 \citep{yang2025qwen3} conduct the mixed multi-task reinforcement learning that learns different domains simultaneously. GLM-4.5 \citep{zeng2025glm} and MiMo-V2-Flash \citep{xiao2026mimo} conduct the separate domain-specific reinforcement learning and then merge models with weight merging \citep{wan2025fusechat,yadav2023ties} or distillation \citep{agarwal2024policy}. However, these representative works do not provide more insights and in-depth analysis about multi-domain RLVR. In this work, we aim to conduct extensive comparison and internal analysis about these two paradigms.

\subsection{Model merging}
There are basically two methodologies for merging multiple domain-specific large language models to a general model which can achieve comparable performance in different domains with the specific model: 1) training-free weight merging \citep{yu2024language} and 2) on/off-policy distillation. By directly blending weights, weight merging achieves functional integration without the high overhead of further training. Beyond the naive average merging, fisher merging \citep{matena2022merging} calculates the fusion weights using the Fisher information matrix; TIES-Merging \citep{yadav2023resolving} resolves task conflicts via pruning, sign agreement and a final disjoint fusion of consistent signs. Besides, we can also distillation the initial model from the multiple domain-specific models. Off-policy distillation conducts supervised fine-tuning using the rollout trajectories generated by multiple domain models. Multi-teacher on-policy distillation \citep{agarwal2024policy} minimizes the Kullback-Leibler divergence between the prediction probability of the student and teacher model on the rollout trajectories, which are generated by the student model.

\section{Experiments and analysis}

\begin{table}
\centering
\scalebox{0.71}{
\begin{tabular}{l|cccccc|c}
\toprule
\textbf{Data Types} & \textbf{Math Formal Proofs} & \textbf{Math} & \textbf{Science} & \textbf{Code} & \textbf{Chat} & \textbf{Conversational Agent} & \textbf{Total} \\
\midrule
$\#$Samples & 335,122 & 2,950,525 & 2,263,340 & 3,927,984 & 4,309,780 & 335,122 & 14,121,873 \\
Proportion ($\%$) & 2.37 & 20.89 & 16.04 & 27.81 & 30.52 & 2.37 & 100.00 \\
\bottomrule
\end{tabular}}
\caption{Our SFT dataset blend strategy. The blend proportion mainly follows the Nemotron 3 Nano technical report.}
\label{sft_blend}
\vspace{-2mm}
\end{table}

\subsection{Preliminary}
Pre-training via next-token prediction equips models with extensive world knowledge, while post-training is the process of learning how to use that knowledge to be a helpful assistant. The post-training phase typically encapsulates multiple stages and mainly contains Supervised Fine-Tuning (SFT) and Reinforcement Learning (RL) \citep{guo2025deepseek,yang2025qwen3}. During the supervised fine-tuning stage, the models are adapted to high-quality, instruction-based datasets to develop basic conversational and task-solving capabilities. During the reinforcement learning stage, the models are optimized based on the reward of their on-policy generated responses. Instead of the reward signals that align with human preferences, the verifiable rewards provide deterministic, gold-standard feedback that eliminates reward hacking and subjective bias. In this work, we apply Group Relative Policy Optimization (GRPO) \citep{shao2024deepseekmath} as our reinforcement learning algorithm.

Multi-domain reinforcement learning is an important topic in the community and its complexity stems from the potential interference between different domains. The existing state-of-the-art models \citep{guo2025deepseek,yang2025qwen3,xiao2026mimo,zeng2025glm} typically apply two different training paradigms: mixed multi-task reinforcement learning and separate reinforcement learning followed by model merging. However, they do not provide much detailed analysis and comparison. In this work, we aim to explore the best ways to combine multi-domain reinforcement learning, and provide detailed comparisons and in-depth analysis to fill the gaps. The following subsections are organized as follows: we first introduce the experimental framework and core results, subsequently diving into a detailed analysis of the underlying mechanisms from multiple perspectives.

\begin{table}[t]
\centering
\scalebox{0.83}{
\begin{tabular}{lccccc|c|c}
\toprule
\textbf{Methods} & \textbf{Math-RL} & \textbf{Coding-RL} & \textbf{Science-RL} & \textbf{IF-RL} & \textbf{Agent-RL} & \textbf{MT-OPD} & \textbf{Multi-Task RL} \\
\midrule
Batch Size & 128 & 128 & 128 & 128 & 128 & 256 & 128 \\
$\#$Rollouts & 16 & 16 & 16 & 16 & 16 & 4 & 16 \\
$\#$Steps & 400 & 400 & 400 & 400 & 400 & 300 & 1000 \\
GPU Hours & 4782.3 & 6404.4 & 1140.8 & 1180.0 & 2271.0 & 1451.9 & 10050.6 \\
\bottomrule
\end{tabular}}
\caption{Training settings and total GPU hours for different RLVR and on-policy distillation training. ``IF'' is instruction following and ``MT-OPD'' is multi-teacher on-policy distillation.}
\label{gpu_hours}
\vspace{-4mm}
\end{table}

\begin{table}[t]
\centering
\scalebox{0.67}{
\begin{tabular}{lcc|cccccc|c}
\toprule
\textbf{Benchmarks} & \textbf{Qwen3-4B-Base} & \textbf{SFT} & \textbf{RL-Math}  & \textbf{RL-Coding} & \textbf{RL-Science} & \textbf{RL-IF} & \textbf{RL-Agent} & \textbf{Merging} & \textbf{RL-Multi} \\ 
\midrule
\multicolumn{10}{c}{\textit{Math Tasks}} \\
\midrule
AIME'24                      & 9.65  & 56.04 & 77.66 & 61.61 & 64.69 & 67.81 & 57.60 & \underline{80.57} & \textbf{81.20} \\
AIME'25                      & 5.68  & 51.30 & 70.42 & 55.16 & 56.35 & 60.52 & 54.64 & \textbf{74.48} & \underline{73.39} \\
\midrule
\multicolumn{10}{c}{\textit{Coding Tasks}} \\
\midrule
LCB v5                       & 16.50 & 55.92 & 59.59 & \textbf{65.00} & 57.66 & 60.80 & 62.95 & \underline{63.31} & 63.21 \\
LCB v6                       & 18.29 & 52.00 & 54.57 & \textbf{58.86} & 53.14 & 53.14 & 54.29 & \underline{58.86} & 56.57 \\
\midrule
\multicolumn{10}{c}{\textit{Science Tasks}} \\
\midrule
HLE                          & 4.45  &  5.75 &  6.39 &  5.92 &  6.26 &  7.04 &  5.98 &  \textbf{7.25} & \underline{7.18} \\
GPQA-Diamond                 & 20.08 & 42.68 & \textbf{58.46} & 46.59 & 53.79 & 49.62 & 46.09 & \underline{55.18} & 53.62 \\
\midrule
\multicolumn{10}{c}{\textit{Instruction Following Tasks}} \\
\midrule
IFEval{\tiny{strict prompt}} & 35.12 & 79.48 & 80.59 & 78.74 & 78.93 & 90.94 & 81.33 & \textbf{93.90} & \underline{93.53} \\
IFBench                      & 11.90 & 38.44 & 40.14 & 39.46 & 40.14 & \underline{59.52} & 40.82 & 59.18 & \textbf{61.22} \\
\midrule
\multicolumn{10}{c}{\textit{Agent Tasks}} \\
\midrule
BFCL v3                      & 29.73 & 50.05 & 50.32 & 50.64 & 49.56 & 50.55 & 59.14 & \textbf{61.37} & \underline{60.74} \\
\bottomrule
\end{tabular}}
\caption{The evaluation scores on 9 benchmarks across 5 different domains. The highest and second-best scores are shown in bold and underlined respectively.}
\label{main_res}
\vspace{-2mm}
\end{table}

\subsection{Experimental design and results}
Although the post-training process in existing works typically involves multiple alternating stages of SFT and RL, we adopt a simplified SFT then RL pipeline for controllable experiments. We mainly focus on five common domains: math, coding, science, instruction following and agent, which contain the complex reasoning, alignment and tool use. We apply the widely-used Qwen3-4B-Base \citep{yang2025qwen3} model as the starting point for supervised fine-tuning to balance both credibility and operability.

\paragraph{Dataset blend.}
We use the open-source datasets from Nemotron 3 Nano \citep{blakeman2025nemotron} for both SFT and RL. The SFT dataset blend strategy is shown in Table \ref{sft_blend}. For RLVR, the dataset of separate domains are as follows: (1) \textit{Math}: 22,056 samples from DAPO~\citep{yu2025dapo} and Skyworks~\citep{he2025skywork,skywork-or1-2025}; (2) \textit{Coding}: 19,169 samples from CodeContests~\citep{li2022competition} and Open-R1~\citep{penedo2025codeforces}; (3) \textit{Science}: 19,670 samples from OpenScienceReasoning-2~\citep{open_science_reasoning_2_2025}; (4) \textit{Instruction Following}: 16,575 samples from WildChat-1M~\citep{zhao2024wildchat} with instructions from Open-Instruct~\citep{lambert2024tulu}; (5) \textit{Agent}: 10,229 samples from Nemotron-RL-agent-workplace-assistant~\citep{blakeman2025nemotron}. The multi-task RLVR uses the directly mixed of these datasets.

\begin{table}[t]
\centering
\scalebox{0.71}{
\begin{tabular}{lccccccccc|c}
\toprule
\textbf{Methods} & \textbf{AIME'24} & \textbf{AIME'25} & \textbf{LCB v5} & \textbf{LCB v6} & \textbf{HLE} & \textbf{GPQA-D} & \textbf{IFEval{\tiny{strict prompt}}} & \textbf{IFBench} & \textbf{BFCL v3} & \textbf{Avg.} \\ 
\midrule
Average   & 67.55 & 64.38 & 58.92 & 57.71 & 6.16 & 50.13 & 84.84 & 40.82 & 53.17 & 53.74 \\
SCE       & 80.73 & 74.84 & 62.01 & 56.57 & 7.28 & 56.57 & 93.35 & 53.74 & 61.19 & 60.70 \\
Ties      & \textbf{81.15} & 74.74 & 60.84 & 57.71 & \textbf{7.92} & \textbf{57.58} & 92.61 & 54.76 & \textbf{61.73} & 61.00 \\
Ties+DARE & 79.84 & 75.89 & 60.71 & 57.71 & 7.41 & 57.07 & 90.94 & 56.12 & 61.05 & 60.75 \\
TA        & 80.47 & \textbf{76.61} & 57.97 & 52.57 & 6.39 & 54.04 & 93.16 & 61.56 & 59.25 & 60.22 \\
TA+DARE   & 81.09 & 76.18 & 58.65 & 56.00 & 6.39 & 55.56 & 93.72 & \textbf{62.59} & 58.76 & 60.99 \\
\midrule
MT-OPD    & 80.57 & 74.48 & 63.31 & \textbf{58.86} & 7.25 & 55.18 & \textbf{93.90} & 59.18 & 61.37 & \textbf{61.57} \\
MT-OFF    & 80.78 & 76.30 & \textbf{63.98}  & 57.14 & 7.23 & 49.24 & 89.09 & 60.88 & 60.12 & 60.53 \\
\bottomrule
\end{tabular}}
\caption{Comparison among different model merging methods and the best result is in bold. ``TA'' denotes task arithmetic merging, ``MT-OPD'' denotes multi-teacher on-policy distillation, and ``MT-OFF'' denotes multi-teacher off-policy distillation. ``LCB'' denotes LiveCodeBench and ``GPQA-D'' denotes GPQA-Diamond.}
\label{weight_merge}
\vspace{-5mm}
\end{table}

\paragraph{Training.}
For SFT, we fine-tune the base model using 14M samples for one epoch. For reinforcement learning, the important training settings and GPU hours are provided in Table \ref{gpu_hours}. Note that, due to the differences in response length and reward calculation cost, the GPU hours of each training step are different among these different domains. More details about reward design and training settings are provided in appendix \ref{app:rl}.

\paragraph{Model merging.}
For merging five domain-specific models, one direct paradigm is weight merging, which blends the parameters of different models with the same structure. Concretely, we use average merging \citep{wortsman2022model}, task arithmetic merging \citep{ilharco2022editing}, Ties-merging \citep{yadav2023ties} and SCE \citep{wan2025fusechat} and also combine them with DARE \citep{yu2024language}. Here we use the supervised fine-tuned model as the anchor model and set the mask ratio as 0.8 for DARE. Besides, another model merging paradigm is using multi-teacher off/on-policy distillation. Concretely, we use the supervised fine-tuned model as the student model and distill it from the routed teachers in five domains. The training prompts are the same as the multi-task RLVR. The training settings and GPU hours are also provided in Table \ref{gpu_hours}, and more details are provided in appendix \ref{app:opd}.

\begin{figure*}[t]
    \centering
    \subcaptionbox{AIME'24\label{acc_step:sub1}}[0.19\textwidth]{
        \includegraphics[width=\linewidth]{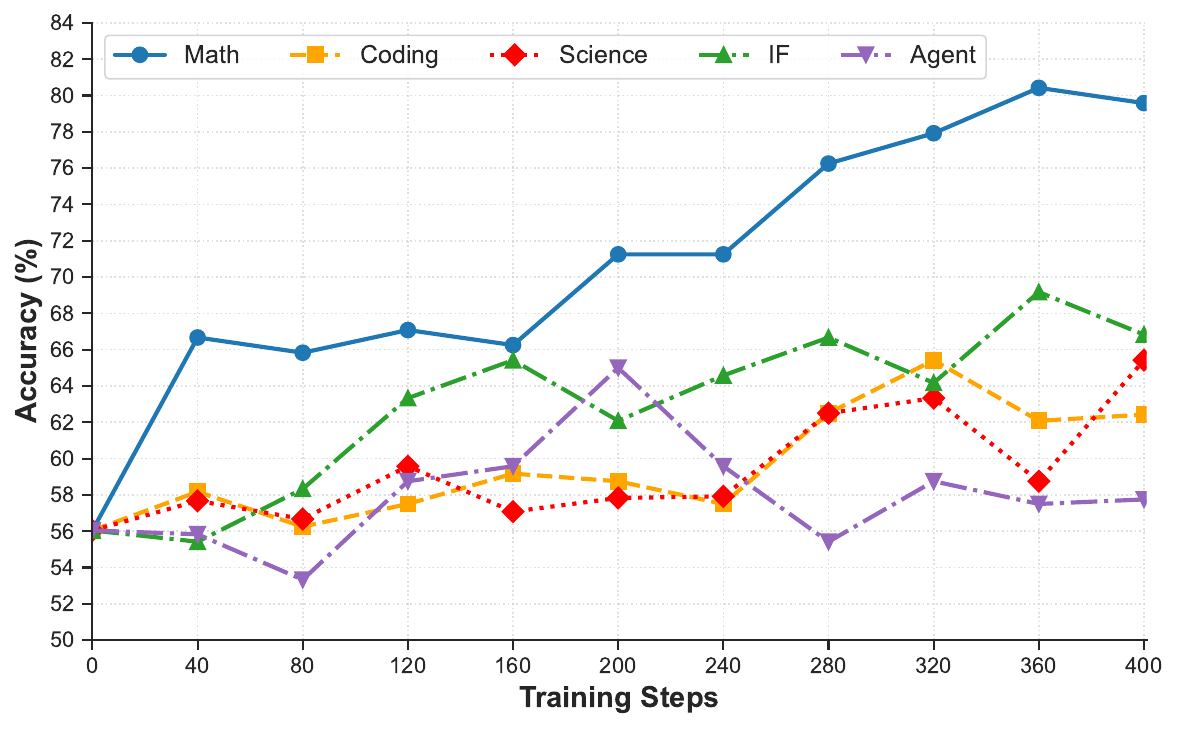}}
    \hfill
    \subcaptionbox{LCB v5\label{acc_step:sub2}}[0.19\textwidth]{
        \includegraphics[width=\linewidth]{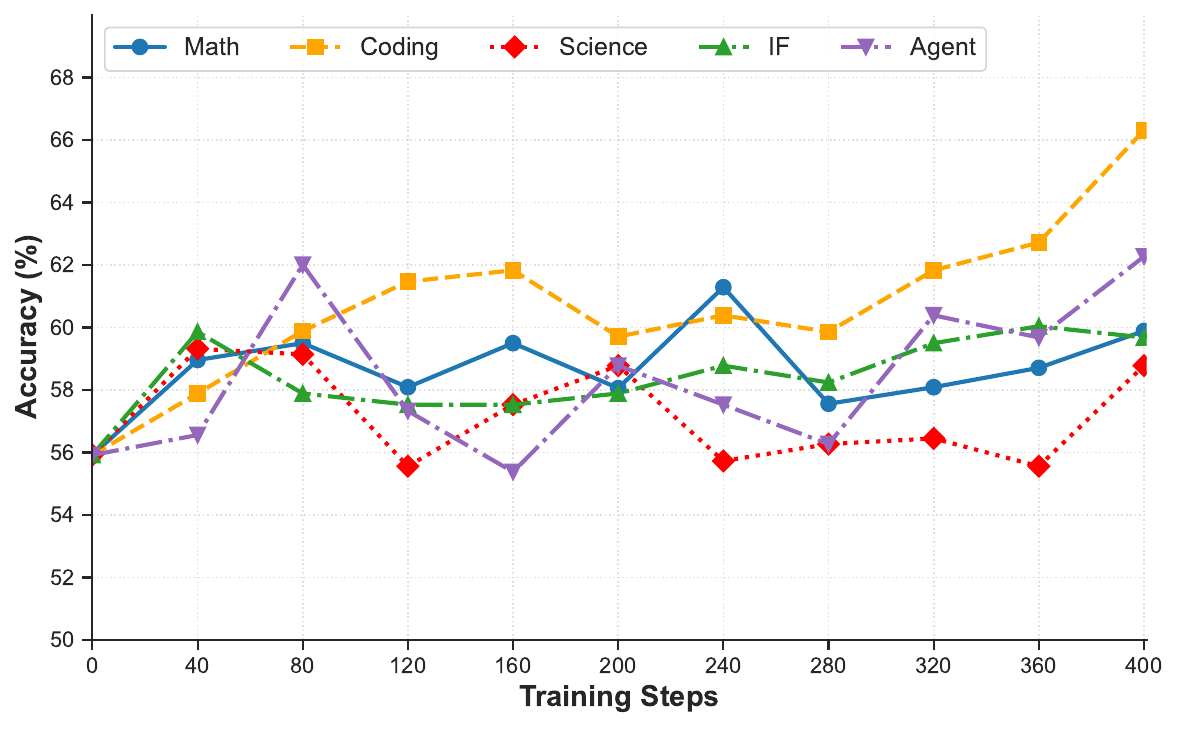}}
    \hfill
    \subcaptionbox{GPQA-D\label{acc_step:sub3}}[0.19\textwidth]{
        \includegraphics[width=\linewidth]{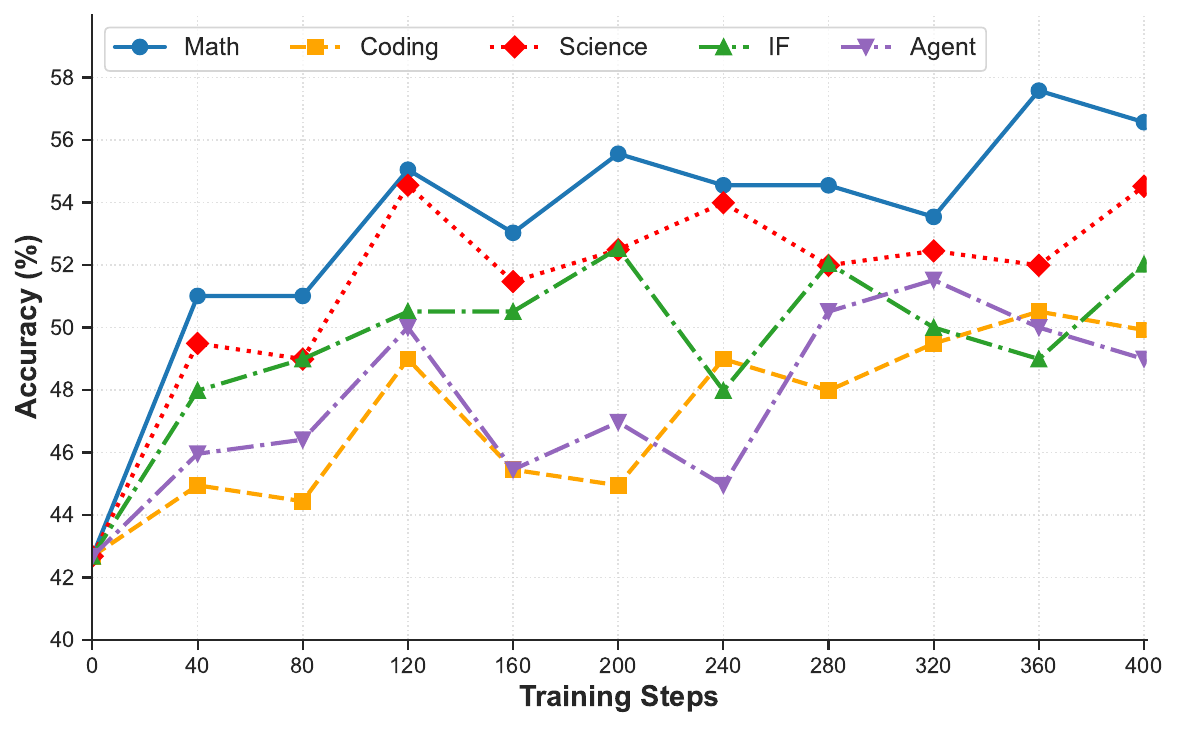}}
    \hfill
    \subcaptionbox{IFEval{\tiny{strict prompt}}\label{acc_step:sub4}}[0.19\textwidth]{
        \includegraphics[width=\linewidth]{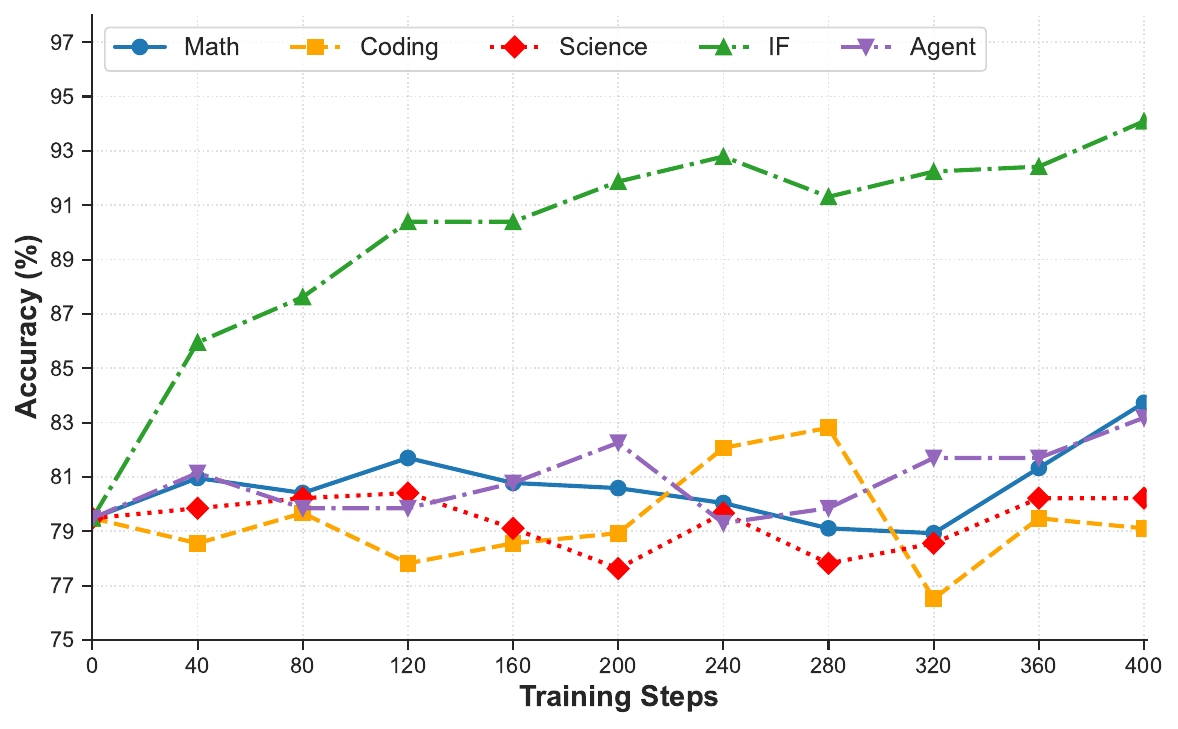}}
    \hfill
    \subcaptionbox{BFCL v3\label{acc_step:sub5}}[0.19\textwidth]{
        \includegraphics[width=\linewidth]{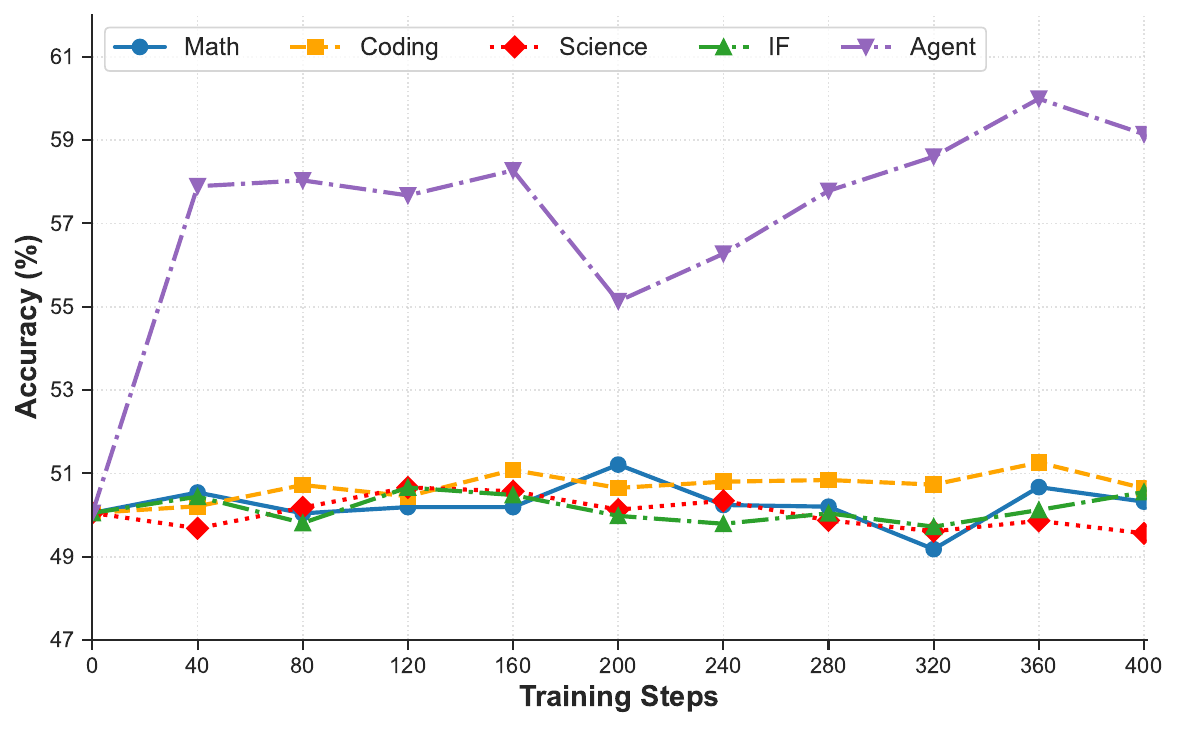}}
    \caption{The accuracy change trajectory of different benchmarks during math, coding, science, instruction following and agent RLVR process.}
    \label{acc_step}
    \vspace{-4mm}
\end{figure*}

\paragraph{Evaluation results.}
The evaluation includes 9 benchmarks: AIME'24 \citep{codeforcesamerican} and AIME'25 \citep{aime} for math tasks, LiveCodeBench v5 and v6 \citep{jain2024livecodebench} for coding tasks, HLE \citep{phan2025humanity} and GPQA-Diamond \citep{rein2024gpqa} for science tasks, IFEval \citep{ifeval} and IFBench \citep{pyatkin2025generalizing} for instruction following tasks, and BFCL v3 \citep{bfcl} for agent tasks. The results are provided in table \ref{main_res}. Firstly, regarding the five distinct RLVR models, the models of math, coding, instruction following and agent domains all achieve the state-of-the-art performance within its respective domain tasks. The RLVR model of math domain obtains better performance than that of science domain at science tasks, and the reason maybe these two science benchmarks require more logical reasoning and numerical calculations than science knowledge. Secondly, the mixed multi-task RLVR can achieve comparable performance with separate RLVR followed by model merging with less GPU hours (i.e., $58.3\%$), which means multi-task RLVR is more efficient. When we use similar computation budget (i.e., five 200-steps single-domain RLVR vs. 1000-steps multi-task RLVR), multi-task RLVR can achieve better performance with average improvement of 4.15 points across 9 benchmarks, and the detailed results are shown in Table \ref{sim_budget}. Thirdly, the gradient interference \citep{yu2020gradient,wu2025imbalanced} among different domains is not significant, and they even have mutual benefits. Concretely, the three reasoning domains (i.e., math, coding and science) can improve each other's performance. The instruction following domain can also improve the performance of these reasoning domains. Regarding the agent domain, when training each domain independently, the other four domains can not improve the agent performance, as shown in Figure \ref{acc_step}. This is because the rollout trajectories of these domains do not involve tool calls or multi-round interactions, so the systematicity of formal logic can not naturally translate into the pragmatic sequences required for tool manipulation. However, this does not imply that the agent domain is fully isolated: when trained jointly, the other domains can still benefit the agent domain. As shown in Table \ref{main_res}, both RL-Multi and Merging outperform the individually trained RL-Agent on BFCL v3, and the pairwise merging results in Table \ref{kl_ablation1} further confirm that combining the agent expert with other domain experts (e.g., Agent+Math, Agent+Coding) consistently improves the agent performance. This suggests that while other domains provide no direct training signal for tool use in isolation, they become beneficial once the agent's own tool-use trajectories are present, and the cross-domain interference on the agent domain remains small. The further evidence is shown in Figure \ref{acc_step}, where we evaluate the accuracy change trajectory throughout the RLVR process of each individual domain and select the benchmarks from various domains for credibility. Fourthly, the comparison among different model merging methods is shown in Table \ref{weight_merge}, where the best results are in bold. Although the multi-teacher on-policy distillation is only slightly better than direct weight merging and requires additional GPU hours, it is still favored by many top-tier works \citep{xiao2026mimo,zeng2025glm}. The reason may be its flexibility and operability and it allows dynamic adjustment of teacher weighting and training schedules during merging. Different model merging methods exhibit a seesaw effect on multiple benchmarks, and we choose the best merging method with the average scores. Model merging not only preserves the most performance of the different domains, but can even achieve further improvements, such as in AIME'24, AIME'25, HLE, IFEval and BFCL v3. This further verifies the gain effect between different domains. Note that our best model using open source dataset achieves comparable performance with official Qwen3-4B model \citep{yang2025qwen3} as shown in Table \ref{off_vs_our}, which verifies the effectiveness of our implementation.

\subsection{Explore policy neighborhoods}

\begin{figure}[t]
    \centering
    \begin{minipage}{0.6\textwidth}
        \centering
        \includegraphics[width=0.6\linewidth]{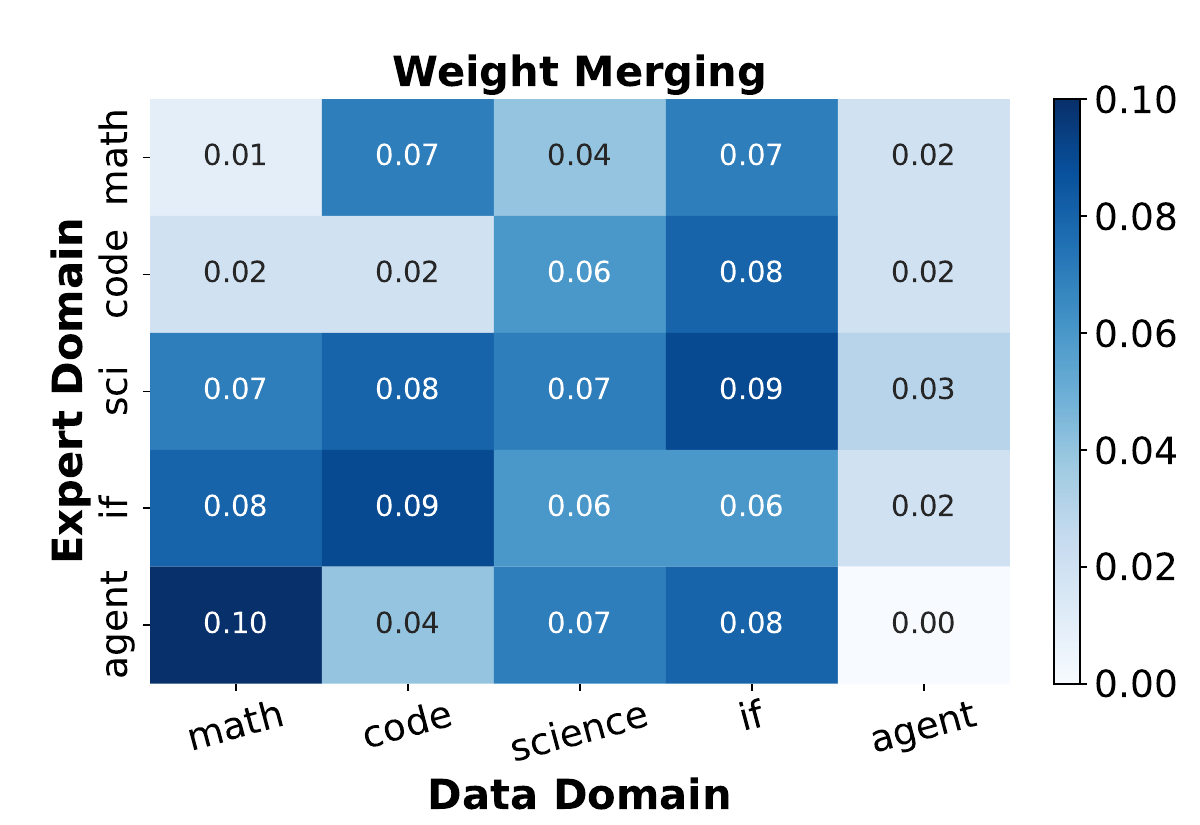}
        \caption{Cross-comparison of KL-divergence (cell value). The y-axis is the domain of the expert model, and the x-axis is the trajectory data domain.}
        \label{fig:kl_heatmap}
    \end{minipage}
    \hfill
    \begin{minipage}{0.35\textwidth}
        \centering
        \includegraphics[width=0.7\linewidth]{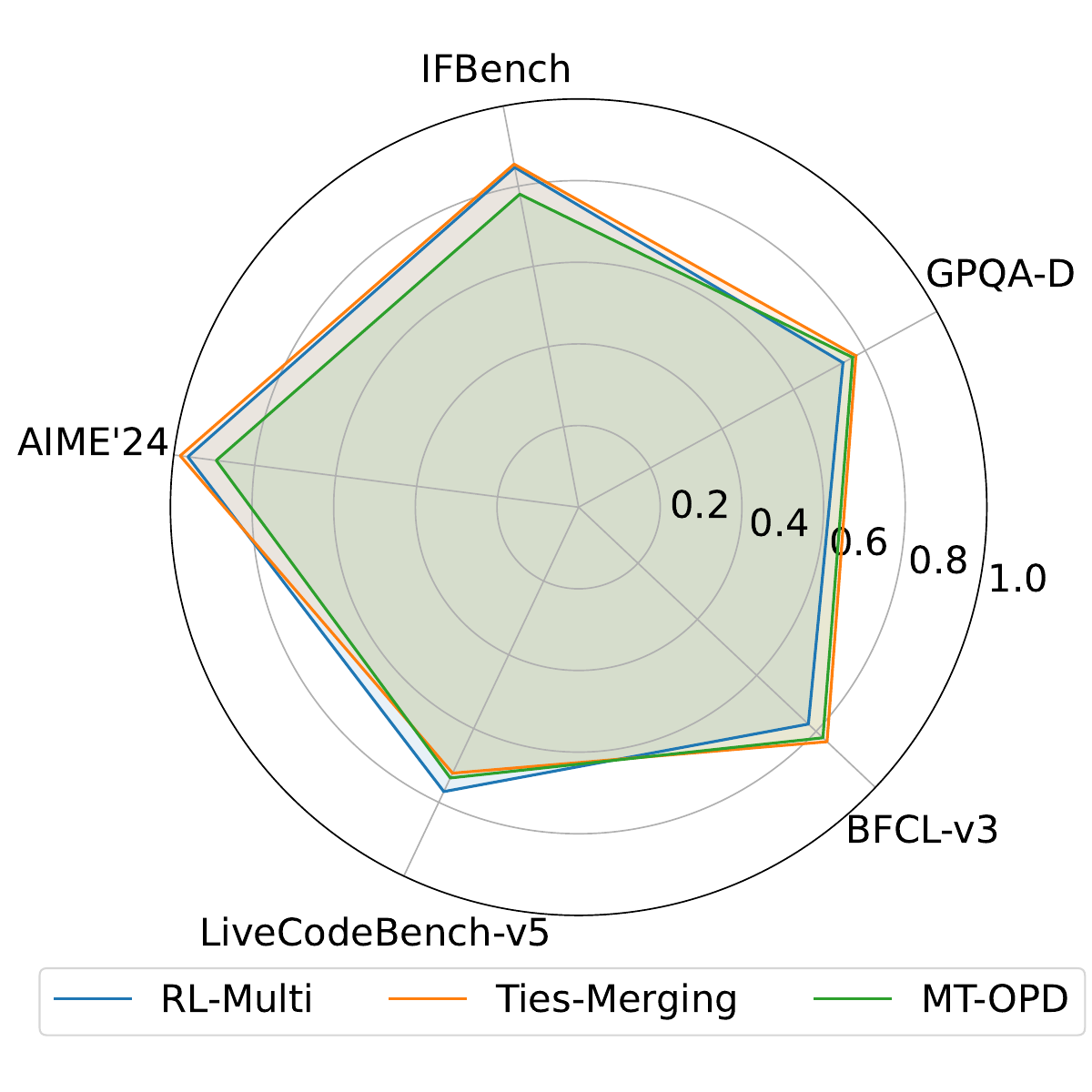}
        \caption{Accuracy gain consistency with union of single-task models on 5 benchmarks.}
        \label{fig:gain-consistency}
    \end{minipage}
    \vspace{-2mm}
\end{figure}

To clearly investigate the reason of performance changes after weight merging, it is necessary to decouple the effects of different domains. Kullback-Leibler (KL) divergence is commonly used to analyze the relationships between LLMs \citep{shenfeld2025rl}. To this end, we cross-compare the KL-divergence between different domain experts and merged multi-domain models on the trajectories across various domains, as shown in Figure \ref{fig:kl_heatmap}. The trajectories are generated by different domain experts. As we can see, for a given data domain, besides corresponding domain expert, the experts from other domains could also exhibit relatively low KL-divergence with the resulting multi-domain policy model. For example, when evaluating on \textit{math} domain, the \textit{math} expert exhibits the lowest KL-divergence with the multi-domain policy model. Remarkably, the \textit{coding} expert also shows a relatively low KL-divergence. Therefore, it is interesting to explore the effects of merging coding expert into the math one. As shown in Table \ref{kl_ablation}, the combined model achieves further performance gains in the \textit{math} domain. These observations imply that, during weight merging, policy distributions from various domains may interact with one another, especially for domain experts whose policies are close to the merged model.

Formally, given a domain $\mathcal{A}$ and its expert model $E_{\mathcal{A}}$,  we define domain $\mathcal{B}$ as a \textbf{policy neighborhood} of $\mathcal{A}$ if the following condition is satisfied:
\begin{equation}
    \mathbb{E}_{x \sim \mathcal{A}, \hat{y} \sim \pi_{E_{\mathcal{B}}} \left( \cdot \mid x\right)}\left[\log \frac{\pi_{E_{\mathcal{B}}} (\hat{y} \mid x)}{\pi_{\text {merge}}(\hat{y} \mid x)}\right] \leq \varepsilon
\end{equation}
where $\pi_{\text {merge}}$ is the merged policy model and $\varepsilon$ is a threshold that should be determined by comparing with $\mathrm{KL}\left(\pi_{E_{\mathcal{A}}} \parallel \pi_{\mathrm{merge}}\right)$. Based on this definition, policy neighborhoods can be identified from Figure \ref{fig:kl_heatmap}, where domain $\mathcal{A}$ and $\mathcal{B}$ can be selected at the x-axis and y-axis, respectively. To detailly verify whether merged multi-domain models benefit from policy neighborhood, we conduct an ablation study on the domain combinations using Ties-merging. The results of reasoning domains are provided in Table \ref{kl_ablation} and those of IF and agent are in Table \ref{kl_ablation1}. If we set $\varepsilon=KL(\pi_{E_{\mathcal{A}}} \parallel \pi_{\mathrm{merge}})+0.02$, an interesting pattern emerges. For a given domain like math, coding and science, merging with the policy neighborhood experts can further improve the domain performance. In contrast, merging non-neighboring policy experts can not yield extra gains. This shows that policy neighborhoods may be one of the factors enabling weight merging to maintain or even enhance the performance of individual domains. Furthermore, we observe that the policy neighborhood relationship is asymmetric. For instance, in \textit{math} domain, the \textit{coding} expert is a neighboring policy for the \textit{math} expert, but not vice versa in \textit{coding} domain. This might be attributed to the inherent asymmetry of the KL divergence.  The policy neighborhoods can be identified from Figure \ref{fig:kl_heatmap} and the corresponding support results are in Table \ref{kl_ablation} and \ref{kl_ablation1}. \textit{Coding} is a neighbor of \textit{math} in \textit{math} domain, and \textit{agent} is a neighbor of \textit{coding} in \textit{coding} domain, all other domains are neighbor of \textit{science} in \textit{science} domain. For IF domain, all other domains are its policy neighborhood domain. For agent domain, math, coding and IF are its policy neighborhood domain. Note that, we discover the policy neighborhood phenomenon, but the choice of $\varepsilon$ maybe relatively complex and not naively $+0.02$, and it requires future exploration.

\begin{table}[t]
\centering
\scalebox{0.69}{
\begin{tabular}{lcc|lcc|lcc}
\toprule
\textbf{Domain(s)} & \textbf{AIME24} & \textbf{AIME25} & \textbf{Domain(s)} & \textbf{LCB v5} & \textbf{LCB v6} & \textbf{Domain(s)} & \textbf{HLE} & \textbf{GPAQ-D} \\ 
\midrule
Math            & 77.66 & 70.42 & Coding & 65.00 & 58.86 & Science & 6.26 & 53.79 \\
\midrule
Math + Coding  & $\text{80.29}_{\textcolor{mygreen}{\uparrow 2.63}}$ & $\text{72.44}_{\textcolor{mygreen}{\uparrow 2.02}}$ & Coding + Math & $\text{60.42}_{\textcolor{red!50}{\downarrow 4.58}}$ & $\text{57.17}_{\textcolor{red!50}{\downarrow 1.69}}$ & Science + Math & $\text{6.26}_{\textcolor{mygreen}{\uparrow 0.00}}$ & $\text{55.94}_{\textcolor{mygreen}{\uparrow 2.15}}$  \\
Math + Science & $\text{75.07}_{\textcolor{red!50}{\downarrow 2.59}}$ & $\text{67.65}_{\textcolor{red!50}{\downarrow 2.77}}$ & Coding + Science & $\text{63.84}_{\textcolor{red!50}{\downarrow 1.16}}$ & $\text{60.00}_{\textcolor{mygreen}{\uparrow 1.14}}$ & Science + Coding & $\text{6.09}_{\textcolor{red!50}{\downarrow 0.17}}$ & $\text{55.10}_{\textcolor{mygreen}{\uparrow 1.31}}$ \\
Math + IF      & $\text{78.11}_{\textcolor{mygreen}{\uparrow 0.45}}$ & $\text{69.04}_{\textcolor{red!50}{\downarrow 1.38}}$ & Coding + IF      & $\text{59.30}_{\textcolor{red!50}{\downarrow 5.70}}$ & $\text{59.43}_{\textcolor{mygreen}{\uparrow 0.57}}$ & Science + IF & $\text{6.19}_{\textcolor{red!50}{\downarrow 0.07}}$ & $\text{55.56}_{\textcolor{mygreen}{\uparrow 1.77}}$  \\
Math + Agent   & $\text{72.94}_{\textcolor{red!50}{\downarrow 4.72}}$ & $\text{63.31}_{\textcolor{red!50}{\downarrow 7.11}}$ & Coding + Agent      & $\text{66.49}_{\textcolor{mygreen}{\uparrow 1.49}}$ & $\text{62.86}_{\textcolor{mygreen}{\uparrow 4.00}}$ & Science + Agent & $\text{5.87}_{\textcolor{red!50}{\downarrow 0.39}}$ & $\text{55.81}_{\textcolor{mygreen}{\uparrow 2.02}}$  \\
\bottomrule
\end{tabular}}
\caption{Performance of models merged from different domain experts using Ties-merging.}
\label{kl_ablation}
\vspace{-4mm}
\end{table}

\begin{table}[t]
\centering
\scalebox{0.8}{
\begin{tabular}{lcc|lc}
\toprule
\textbf{Domain(s)} & \textbf{IFEval} & \textbf{IFBench} & \textbf{Domain(s)} & \textbf{BFCL v3} \\ 
\midrule
IF           & 90.94 & 59.52 & Agent & 59.14 \\
\midrule
IF + Math    & $\text{93.35}_{\textcolor{mygreen}{\uparrow 2.41}}$ & $\text{58.84}_{\textcolor{red!50}{\downarrow 0.68}}$ & Agent + Math    & $\text{61.46}_{\textcolor{mygreen}{\uparrow 2.32}}$ \\
IF + Coding  & $\text{93.90}_{\textcolor{mygreen}{\uparrow 2.96}}$ & $\text{58.84}_{\textcolor{red!50}{\downarrow 0.68}}$ & Agent + Coding  & $\text{62.28}_{\textcolor{mygreen}{\uparrow 3.14}}$ \\
IF + Science & $\text{92.98}_{\textcolor{mygreen}{\uparrow 2.04}}$ & $\text{58.84}_{\textcolor{red!50}{\downarrow 0.68}}$ & Agent + Science & $\text{58.80}_{\textcolor{red!50}{\downarrow 0.34}}$ \\
IF + Agent   & $\text{92.68}_{\textcolor{mygreen}{\uparrow 1.74}}$ & $\text{58.82}_{\textcolor{red!50}{\downarrow 0.70}}$ & Agent + IF      & $\text{60.40}_{\textcolor{mygreen}{\uparrow 1.26}}$ \\
\bottomrule
\end{tabular}}
\caption{Performance of models merged from different domain experts using Ties-merging.}
\label{kl_ablation1}
\end{table}

\subsection{Do multi-task and merging models acquire same skills as single-task models?}
The preceding experiments demonstrate that both multi-task training and model merging effectively develop expertise across multiple domains. A natural question arises: do these multi-domain models acquire the same skills as their single-task counterparts? To investigate this, we analyze the overlap of newly solved instances (relative to the SFT baseline) between multi-domain models and the union of five single-task models. Specifically, for each task \(t\), we define a gain vector $g_m^t=(\max(a_m^t(1)-a_{\rm sft}^t(1),0),\cdots,\max(a_m^t(n_t)-a_{\rm sft}^t(n_t),0))$, where $m$ denotes the model, $n_t$ is the size of task $t$'s test set, and $a_m^t(i)\in\{0,1\}$ represents the accuracy of model $m$ on the $i$-th the sample of task $t$. We then construct a union gain vector $g_{\rm union}^t=\max(g_{\rm math}^t,g_{\rm science}^t,g_{\rm coding}^t,g_{\rm IF}^t,g_{\rm agent}^t)$ to serve as a proxy for the collective skills acquired during the learning of single-task. Finally, we compute the cosine similarity between $g_{\rm union}^t$ and the gain vectors of RL-Multi, Ties-merging, and MT-OPD respectively, as the measure of gain consistency. A higher similarity score indicates that a model inherits a greater proportion of skills originally developed through single-task learning.

As illustrated in Figure~\ref{fig:gain-consistency}, all three multi-domain models exhibit significant overlap with single-task models in their learned capabilities. Among the 5 benchmarks, the math task shows the highest consistency in performance gains, suggesting that mathematical skills may be more homogeneous (inherent) and resistant to inter-task interference. In contrast, for other domains, all three models appear to develop distinct proficiencies that are not captured during single-task learning.

In the cross-method comparison, the gain consistency of the weight merging method (Ties-merging) is higher than that of RL-Multi and MT-OPD on most domains. This observation provides an insight: wight merging primarily inherits the original capabilities of the single-task models, whereas the capabilities learned through multi-task training and on-policy distillation exhibit a larger divergence from those learned via single-task training. It confirms the existence of emergent capabilities in multi-task models, which arise from tasks mutually promoting each other during learning.

\subsection{Dynamics of self-verification}\label{sec:verification_horizon}

\begin{figure}[t]
    \centering
    \begin{minipage}{0.53\textwidth}
        \centering
        \begin{subfigure}{0.485\linewidth}
            \centering
            \includegraphics[width=\linewidth]{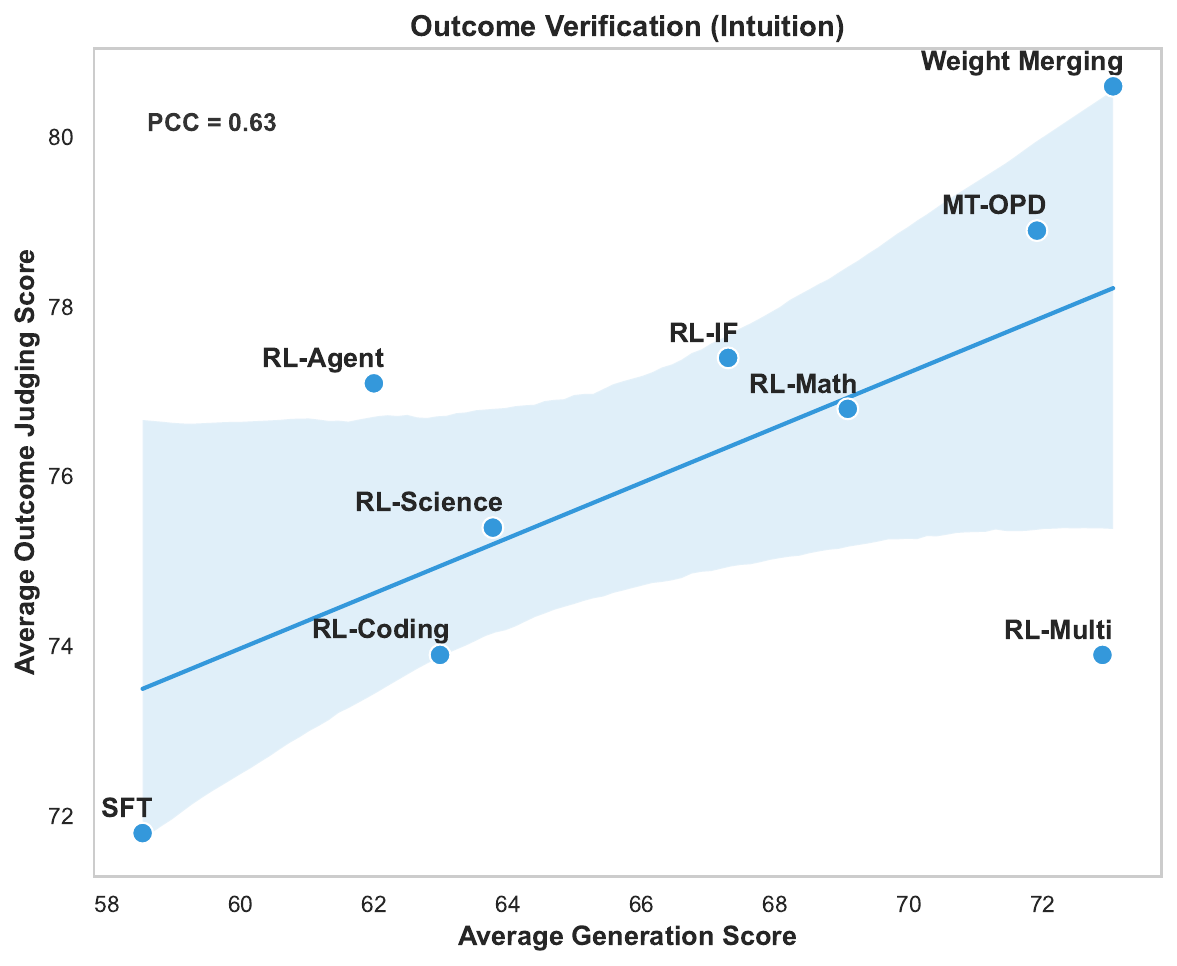}
            \caption{Outcome}
        \end{subfigure}
        \hfill
        \begin{subfigure}{0.485\linewidth}
            \centering
            \includegraphics[width=\linewidth]{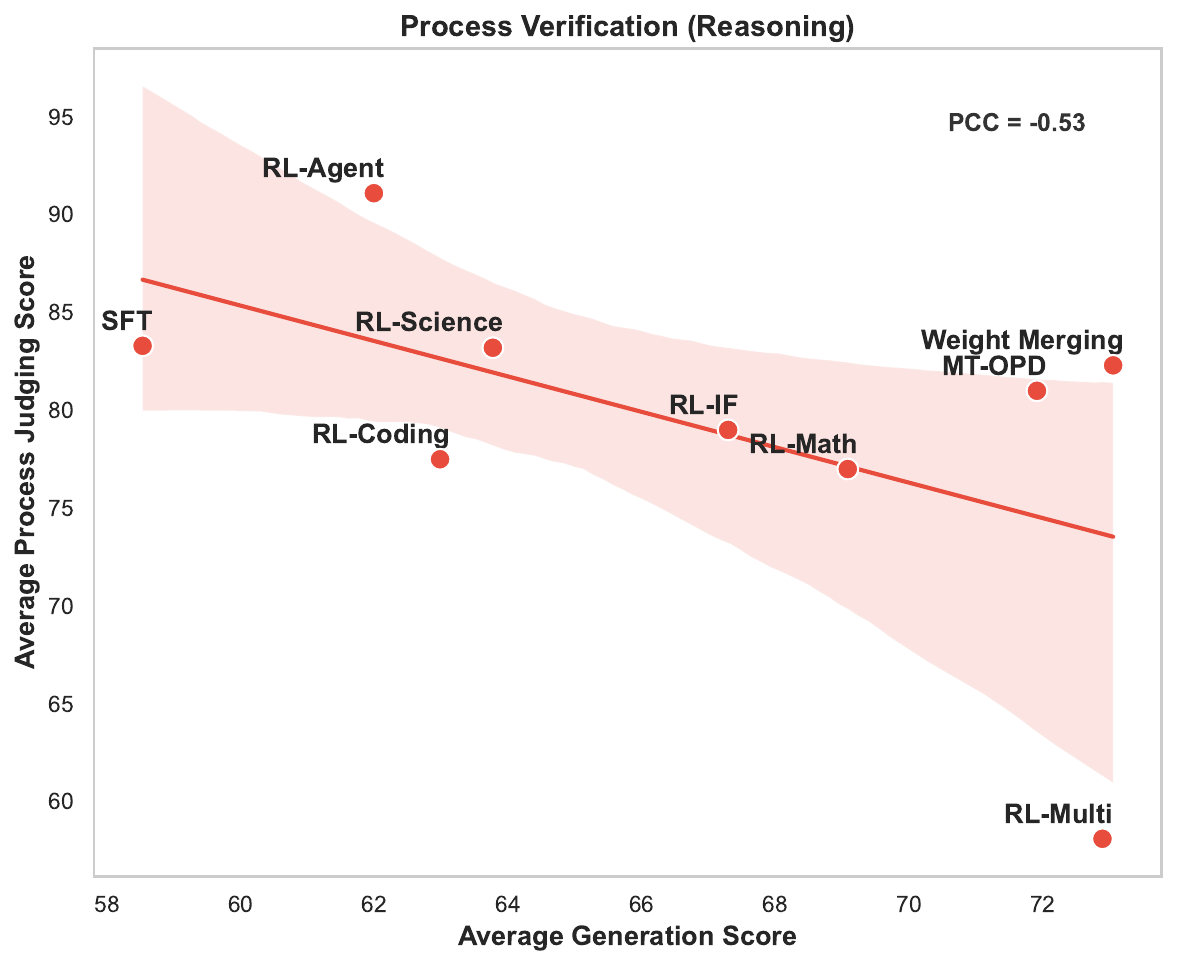}
            \caption{Process}
        \end{subfigure}
        \caption{Correlation between average generation score and average judge score.}
        \label{fig:judge_corr}
    \end{minipage}
    \hfill
    \begin{minipage}{0.45\textwidth}
        \centering
        \includegraphics[width=0.93\linewidth]{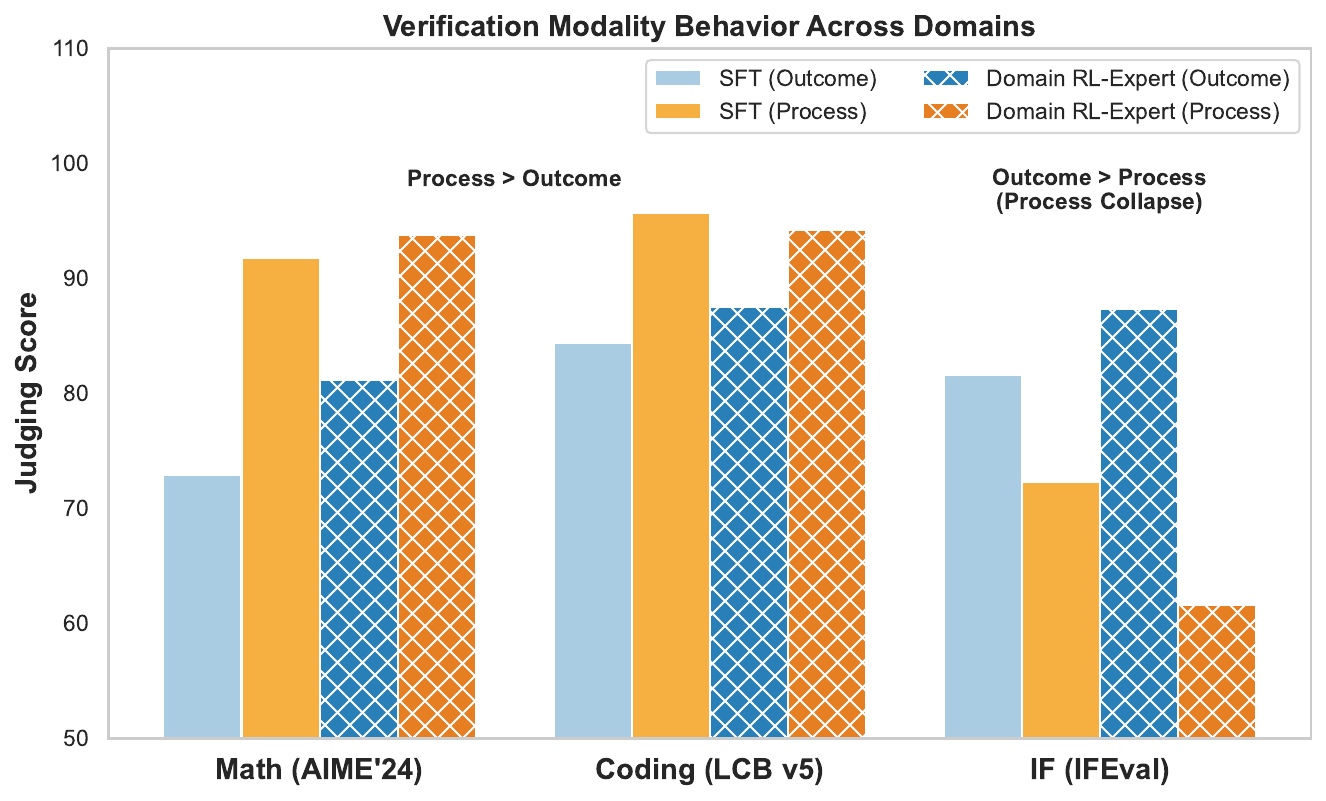}
        \caption{Verification modality behavior across reasoning/constraint domains.}
        \label{fig:verification_horizon}
    \end{minipage}
\vspace{-2mm}
\end{figure}

In this section, we evaluate our RL-trained models as Generative Reward Models (GenRMs) operating on their own trajectories. We use the model itself to verify whether its trajectory can correctly answer the question and further obtain a judge score with the ground-truth answer. We contrast two verification modalities: outcome-based verification where the verifier observes only the final answer (approximating intuition), and process-based verification where the verifier observes the full Chain-of-Thought (approximating reasoning). We calculate the correlation between the expert model's generation score (i.e., evaluation performance) and its judging score in its corresponding domain. The results are provided in Figure~\ref{fig:judge_corr}, where we report the average score across math, coding, science and IF domain using AIME'24, LCB v5, GPQA-D and IFEval benchmarks respectively. We observe a positive correlation (i.e., Pearson Correlation Coefficient (PCC) $r=0.63$) between generation performance and outcome-based judge ability. Conversely, process-based judge ability exhibits a negative correlation (i.e., $r=-0.53$) with generation score, revealing more complex dynamics depending on the task characteristic and training methods. This indicates that while intuitive outcome-level self-verification emerges naturally alongside generation ability improvements, rigorous process-level verification can actually degrade as models optimize for task-specific generation during RLVR.

\paragraph{Finding 1: task characteristic dictates the verification modality.}
The effectiveness of a verification modality is heavily dependent on the inherent structure of the task domain. As shown in Figure \ref{fig:verification_horizon}, for logic-intensive tasks like math and coding, process-based verification consistently outperforms outcome-based verification for both SFT and RL. In these domains, the final answer is a highly compressed representation of a complex derivation. Evaluating only the outcome forces the verifier to make uninformed estimations, whereas process verification enables the model to detect logical fractures step-by-step before they propagate to the final result. Conversely, for constraint-intensive tasks like instruction following, process-based verification severely underperforms. In constraint satisfaction, errors typically manifest in surface-level execution, such as violating JSON syntax or length restrictions. When evaluating these tasks, the reasoning trace often merely states the model's intent (e.g., ``I will output exactly three bullet points''), while the final output reflects the actual execution. Relying on process verification here creates an intent-execution gap: the judge positively evaluates the correct plan within the reasoning trace, resulting in false positives that blind the model to actual formatting failures in the output.

\begin{table}[t]
\centering
\scalebox{0.69}{
\begin{tabular}{l|cc|cc|cc|cc|cc}
\toprule
\multirow{2}{*}{Model} & \multicolumn{2}{c|}{AIME24 (Math)} & \multicolumn{2}{c|}{IFEval (Inst)} & \multicolumn{2}{c|}{LCB v5 (Code)} & \multicolumn{2}{c|}{GPQA (Science)} & \multicolumn{2}{c}{Avg} \\
 & Outcome & Process & Outcome & Process & Outcome & Process & Outcome & Process & Outcome & Process \\
\midrule
SFT             &  72.9 & 91.8 &  81.6 & 72.3 &  84.4 & 95.7 &  48.3 & 73.4 &  71.8 & 83.3 \\
\midrule
RL-Math         &  81.2 & 93.8 &  77.8 & 64.2 & 87.2 & 93.7 &  \textbf{60.9} & 56.2 &  76.8 & 77.0 \\
RL-Coding       &  75.8 & 82.9 & 81.0 & 65.5 &   87.5 & 94.2 &  51.2 & 67.5 &  73.9 & 77.5 \\
RL-Science      &   67.1 & 86.2 &  84.6 & 71.0 & 90.9 & 97.1 &  59.1 & 78.5 &  75.4 & 83.2 \\
RL-IF           &   83.8 & 90.4 &  87.4 & 61.6 &  82.4 & 91.6 &  56.0 & 72.3 &  77.4 & 79.0 \\
RL-Agent        &   76.7 & \textbf{95.3}  & 85.1 & \textbf{88.5} &  \textbf{91.6} & \textbf{99.4} & 55.1 & \textbf{81.1} &  77.1 & \textbf{91.1} \\
Weight Merging  &   82.9 & 93.3 &   \textbf{91.0} & 79.1 &  88.4 & 95.8 & 59.9 & 61.1 &  \textbf{80.6} & 82.3 \\
MT-OPD          &  85.8 & 91.2 & 87.5 & 66.5 &  85.0 & 94.7 & 57.1 & 71.4 & 78.9 & 81.0 \\
\midrule
RL-Multi        &   \textbf{86.7} & 86.2 &   80.5 & 27.5 &  72.3 & 76.1 &  55.9 & 42.6 &  73.9 & 58.1 \\
\bottomrule
\end{tabular}}
\caption{Outcome / process based judge scores of different models on 4 benchmarks.}
\label{tab:generation_critic_comparison}
\vspace{-4mm}
\end{table}

\paragraph{Finding 2: the agent expert excels at process verification.}
We observe that the RL-Agent, optimized specifically for multi-turn general tool-use tasks, significantly outperforms other single-domain models in process-based verification. As shown in Table~\ref{tab:generation_critic_comparison}, the RL-Agent achieves the highest process judging scores across diverse domains, including 95.3 on AIME, 88.5 on IFEval, 99.4 on LCB, and 81.1 on GPQA. Unlike static mathematical derivations, agentic training involves state interactions where the model must continuously evaluate intermediate tool-call returns and environmental feedback. This multi-step optimization forces the model to verify its own trajectory dynamically. Consequently, the model develops a highly robust sequential monitoring capability. This indicates that interactive, multi-turn training is a critical catalyst for cultivating reliable process judges, effectively training the model to treat the reasoning trajectories as functional, verifiable logs.

\paragraph{Finding 3: robustness-efficient trade-off between multi-task RL and model merging.}
While combining diverse training signals theoretically yields comprehensive models, the multi-task training shows a robustness-efficient trade-off phenomenon. RL-Multi achieves comparable performance with model merging paradigm using less GPU hours, but it also induces a severe degradation in process verification capabilities. Although the performance is comparable as shown in Table \ref{main_res}, the average process verification score of RL-Multi collapses to 58.1 as shown in Table \ref{tab:generation_critic_comparison}, far behind model merging methods. This phenomenon is a signal of some kind of interference between different domains during multi-task RLVR training. The model's internal verification misaligns, optimizing for superficial text patterns rather than rigorous evaluation. In contrast, methods that decouple expert training before merging are significantly more stable. Weight merging, which operates directly in the parameter space by averaging the weights of expert models, achieves the highest average outcome judgment (80.6). This demonstrates that weight merging acts as an effective regularizer, preserving general capabilities while avoiding the over-optimized noise like in multi-task training. Meanwhile, MT-OPD integrates expert knowledge in the behavior space. By utilizing multi-teacher supervisions on generated trajectories, MT-OPD prevents the model from overfitting to a single domain's reasoning style, securing a robust and balanced foundation for both outcome and process based verification.

\subsection{Discussion about guidance and inspiration}
The above findings yield actionable guidelines for developing multi-domain or single-domain expert models through RLVR:
\begin{itemize}
\item Multi-task RLVR achieves comparable performance with separate RLVR followed by model merging with less GPU Hours. Therefore, we recommend multi-task RLVR training when computational resources are limited or emergent cross-domain capabilities are desired. But when existing expert models are available, maximum single-domain performance is critical or self-verification ability is crucial, separate RLVR training followed by model merging is preferred.
\item The mutual gain between reasoning domains during multi-task training inspires us that we can use the data from beneficial domains to enhance our target domain and obtain a better single-domain expert model. For example, if we want to obtain a expert-level math LLM, using some coding and STEM data would be very helpful, so as the instruction following data. The policy neighborhood phenomenon also reveals the potential for merging other experts to enhance the target expert.
\item Section \ref{sec:verification_horizon} reveals a critical alignment between verification modality and error locus, which yields concrete design principles for reward functions. For logic-intensive tasks (e.g., math or coding), errors are hidden in the process, so process-based verification is essential. For constraint-intensive tasks (e.g., instruction following), errors manifest in execution failures and outcome-based verification is superior.
\item Training paradigm selection can align with capability acquisition objectives: model merging primarily inherits existing single-task capabilities with high predictability, suitable for aggregating established expertise, whereas multi-task training develops emergent problem-solving strategies through cross-domain synergy, advantageous for cultivating generalist models with novel competencies beyond skill aggregation.
\end{itemize}

\section{Conclusion}
In this work, we present a systematic comparison between mixed multi-task RLVR and separate domain-specific RLVR followed by model merging across five diverse domains. Our findings reveal that multi-task RLVR exhibits small inter-task interference and demonstrates significant synergistic effects among reasoning-intensive domains, driven by neighborhood policy transfer mechanisms. We identify distinct capability acquisition patterns: weight merging primarily inherits single-task capabilities, while multi-task training develops emergent problem-solving strategies. Furthermore, we uncover that RLVR induces task-sensitive self-verification capabilities, with a critical robustness-efficiency trade-off between training paradigms—multi-task RLVR achieves comparable performance with $58.3\%$ computational cost but at the expense of process-level verification, whereas model merging preserves robust verification through decoupled expert integration. These insights provide actionable guidance for developing general reasoning models.

\bibliography{colm2026_conference}

@article{guo2025deepseek,
  title={Deepseek-r1: Incentivizing reasoning capability in llms via reinforcement learning},
  author={Guo, Daya and Yang, Dejian and Zhang, Haowei and Song, Junxiao and Zhang, Ruoyu and Xu, Runxin and Zhu, Qihao and Ma, Shirong and Wang, Peiyi and Bi, Xiao and others},
  journal={arXiv preprint arXiv:2501.12948},
  year={2025}
}

@article{zhang2025survey,
  title={A survey of reinforcement learning for large reasoning models},
  author={Zhang, Kaiyan and Zuo, Yuxin and He, Bingxiang and Sun, Youbang and Liu, Runze and Jiang, Che and Fan, Yuchen and Tian, Kai and Jia, Guoli and Li, Pengfei and others},
  journal={arXiv preprint arXiv:2509.08827},
  year={2025}
}

@article{yue2025does,
  title={Does reinforcement learning really incentivize reasoning capacity in llms beyond the base model?},
  author={Yue, Yang and Chen, Zhiqi and Lu, Rui and Zhao, Andrew and Wang, Zhaokai and Song, Shiji and Huang, Gao},
  journal={arXiv preprint arXiv:2504.13837},
  year={2025}
}

@article{albalak2025big,
  title={Big-math: A large-scale, high-quality math dataset for reinforcement learning in language models},
  author={Albalak, Alon and Phung, Duy and Lile, Nathan and Rafailov, Rafael and Gandhi, Kanishk and Castricato, Louis and Singh, Anikait and Blagden, Chase and Xiang, Violet and Mahan, Dakota and others},
  journal={arXiv preprint arXiv:2502.17387},
  year={2025}
}

@article{chen2025r1,
  title={R1-Code-Interpreter: Training LLMs to Reason with Code via Supervised and Reinforcement Learning},
  author={Chen, Yongchao and Liu, Yueying and Zhou, Junwei and Hao, Yilun and Wang, Jingquan and Zhang, Yang and Fan, Chuchu},
  journal={arXiv preprint arXiv:2505.21668},
  year={2025}
}

@article{lambert2024tulu,
  title={Tulu 3: Pushing frontiers in open language model post-training},
  author={Lambert, Nathan and Morrison, Jacob and Pyatkin, Valentina and Huang, Shengyi and Ivison, Hamish and Brahman, Faeze and Miranda, Lester James V and Liu, Alisa and Dziri, Nouha and Lyu, Shane and others},
  journal={arXiv preprint arXiv:2411.15124},
  year={2024}
}

@article{shao2024deepseekmath,
  title={Deepseekmath: Pushing the limits of mathematical reasoning in open language models},
  author={Shao, Zhihong and Wang, Peiyi and Zhu, Qihao and Xu, Runxin and Song, Junxiao and Bi, Xiao and Zhang, Haowei and Zhang, Mingchuan and Li, YK and Wu, Yang and others},
  journal={arXiv preprint arXiv:2402.03300},
  year={2024}
}

@article{zheng2025group,
  title={Group sequence policy optimization},
  author={Zheng, Chujie and Liu, Shixuan and Li, Mingze and Chen, Xiong-Hui and Yu, Bowen and Gao, Chang and Dang, Kai and Liu, Yuqiong and Men, Rui and Yang, An and others},
  journal={arXiv preprint arXiv:2507.18071},
  year={2025}
}

@article{yu2025dapo,
  title={Dapo: An open-source llm reinforcement learning system at scale},
  author={Yu, Qiying and Zhang, Zheng and Zhu, Ruofei and Yuan, Yufeng and Zuo, Xiaochen and Yue, Yu and Dai, Weinan and Fan, Tiantian and Liu, Gaohong and Liu, Lingjun and others},
  journal={arXiv preprint arXiv:2503.14476},
  year={2025}
}

@article{cui2025process,
  title={Process reinforcement through implicit rewards},
  author={Cui, Ganqu and Yuan, Lifan and Wang, Zefan and Wang, Hanbin and Zhang, Yuchen and Chen, Jiacheng and Li, Wendi and He, Bingxiang and Fan, Yuchen and Yu, Tianyu and others},
  journal={arXiv preprint arXiv:2502.01456},
  year={2025}
}

@article{dong2025agentic,
  title={Agentic reinforced policy optimization},
  author={Dong, Guanting and Mao, Hangyu and Ma, Kai and Bao, Licheng and Chen, Yifei and Wang, Zhongyuan and Chen, Zhongxia and Du, Jiazhen and Wang, Huiyang and Zhang, Fuzheng and others},
  journal={arXiv preprint arXiv:2507.19849},
  year={2025}
}

@article{zhu2024deepseek,
  title={Deepseek-coder-v2: Breaking the barrier of closed-source models in code intelligence},
  author={Zhu, Qihao and Guo, Daya and Shao, Zhihong and Yang, Dejian and Wang, Peiyi and Xu, Runxin and Wu, Y and Li, Yukun and Gao, Huazuo and Ma, Shirong and others},
  journal={arXiv preprint arXiv:2406.11931},
  year={2024}
}

@article{hui2024qwen2,
  title={Qwen2. 5-coder technical report},
  author={Hui, Binyuan and Yang, Jian and Cui, Zeyu and Yang, Jiaxi and Liu, Dayiheng and Zhang, Lei and Liu, Tianyu and Zhang, Jiajun and Yu, Bowen and Lu, Keming and others},
  journal={arXiv preprint arXiv:2409.12186},
  year={2024}
}

@article{yang2024qwen2,
  title={Qwen2. 5-math technical report: Toward mathematical expert model via self-improvement},
  author={Yang, An and Zhang, Beichen and Hui, Binyuan and Gao, Bofei and Yu, Bowen and Li, Chengpeng and Liu, Dayiheng and Tu, Jianhong and Zhou, Jingren and Lin, Junyang and others},
  journal={arXiv preprint arXiv:2409.12122},
  year={2024}
}

@article{yang2025qwen3,
  title={Qwen3 technical report},
  author={Yang, An and Li, Anfeng and Yang, Baosong and Zhang, Beichen and Hui, Binyuan and Zheng, Bo and Yu, Bowen and Gao, Chang and Huang, Chengen and Lv, Chenxu and others},
  journal={arXiv preprint arXiv:2505.09388},
  year={2025}
}

@article{zeng2025glm,
  title={Glm-4.5: Agentic, reasoning, and coding (arc) foundation models},
  author={Zeng, Aohan and Lv, Xin and Zheng, Qinkai and Hou, Zhenyu and Chen, Bin and Xie, Chengxing and Wang, Cunxiang and Yin, Da and Zeng, Hao and Zhang, Jiajie and others},
  journal={arXiv preprint arXiv:2508.06471},
  year={2025}
}

@article{xiao2026mimo,
  title={MiMo-V2-Flash Technical Report},
  author={Xiao, Bangjun and Xia, Bingquan and Yang, Bo and Gao, Bofei and Shen, Bowen and Zhang, Chen and He, Chenhong and Lou, Chiheng and Luo, Fuli and Wang, Gang and others},
  journal={arXiv preprint arXiv:2601.02780},
  year={2026}
}

@inproceedings{wan2025fusechat,
  title={Fusechat: Knowledge fusion of chat models},
  author={Wan, Fanqi and Zhong, Longguang and Yang, Ziyi and Chen, Ruijun and Quan, Xiaojun},
  booktitle={Proceedings of the 2025 Conference on Empirical Methods in Natural Language Processing},
  pages={21629--21653},
  year={2025}
}

@inproceedings{agarwal2024policy,
  title={On-policy distillation of language models: Learning from self-generated mistakes},
  author={Agarwal, Rishabh and Vieillard, Nino and Zhou, Yongchao and Stanczyk, Piotr and Garea, Sabela Ramos and Geist, Matthieu and Bachem, Olivier},
  booktitle={The twelfth international conference on learning representations},
  year={2024}
}

@article{yadav2023ties,
  title={Ties-merging: Resolving interference when merging models},
  author={Yadav, Prateek and Tam, Derek and Choshen, Leshem and Raffel, Colin A and Bansal, Mohit},
  journal={Advances in Neural Information Processing Systems},
  volume={36},
  pages={7093--7115},
  year={2023}
}

@inproceedings{yu2024language,
  title={Language models are super mario: Absorbing abilities from homologous models as a free lunch},
  author={Yu, Le and Yu, Bowen and Yu, Haiyang and Huang, Fei and Li, Yongbin},
  booktitle={Forty-first International Conference on Machine Learning},
  year={2024}
}

@article{matena2022merging,
  title={Merging models with fisher-weighted averaging},
  author={Matena, Michael S and Raffel, Colin A},
  journal={Advances in Neural Information Processing Systems},
  volume={35},
  pages={17703--17716},
  year={2022}
}

@article{yadav2023resolving,
  title={Resolving interference when merging models},
  author={Yadav, Prateek and Tam, Derek and Choshen, Leshem and Raffel, Colin and Bansal, Mohit},
  journal={arXiv preprint arXiv:2306.01708},
  volume={1},
  year={2023}
}

@article{jaech2024openai,
  title={Openai o1 system card},
  author={Jaech, Aaron and Kalai, Adam and Lerer, Adam and Richardson, Adam and El-Kishky, Ahmed and Low, Aiden and Helyar, Alec and Madry, Aleksander and Beutel, Alex and Carney, Alex and others},
  journal={arXiv preprint arXiv:2412.16720},
  year={2024}
}

@article{wen2025reinforcement,
  title={Reinforcement learning with verifiable rewards implicitly incentivizes correct reasoning in base llms},
  author={Wen, Xumeng and Liu, Zihan and Zheng, Shun and Ye, Shengyu and Wu, Zhirong and Wang, Yang and Xu, Zhijian and Liang, Xiao and Li, Junjie and Miao, Ziming and others},
  journal={arXiv preprint arXiv:2506.14245},
  year={2025}
}

@article{gao2025soft,
  title={Soft adaptive policy optimization},
  author={Gao, Chang and Zheng, Chujie and Chen, Xiong-Hui and Dang, Kai and Liu, Shixuan and Yu, Bowen and Yang, An and Bai, Shuai and Zhou, Jingren and Lin, Junyang},
  journal={arXiv preprint arXiv:2511.20347},
  year={2025}
}

@article{jain2024livecodebench,
  title={Livecodebench: Holistic and contamination free evaluation of large language models for code},
  author={Jain, Naman and Han, King and Gu, Alex and Li, Wen-Ding and Yan, Fanjia and Zhang, Tianjun and Wang, Sida and Solar-Lezama, Armando and Sen, Koushik and Stoica, Ion},
  journal={arXiv preprint arXiv:2403.07974},
  year={2024}
}

@misc{aime,
      title={{AIME} Problems and Solutions},
      author={{AIME}},
      year={2025},
      url={https://artofproblemsolving.com/wiki/index.php/AIME_Problems_and_Solutions}
}

@misc{codeforcesamerican,
  title={American Invitational Mathematics Examination-AIME 2024, 2024},
  author={MAA},
  year={2024}
}

@article{ifeval,
  author       = {Jeffrey Zhou and
                  Tianjian Lu and
                  Swaroop Mishra and
                  Siddhartha Brahma and
                  Sujoy Basu and
                  Yi Luan and
                  Denny Zhou and
                  Le Hou},
  title        = {Instruction-Following Evaluation for Large Language Models},
  journal      = {CoRR},
  volume       = {abs/2311.07911},
  year         = {2023}
}

@article{phan2025humanity,
  title={Humanity's last exam},
  author={Phan, Long and Gatti, Alice and Han, Ziwen and Li, Nathaniel and Hu, Josephina and Zhang, Hugh and Zhang, Chen Bo Calvin and Shaaban, Mohamed and Ling, John and Shi, Sean and others},
  journal={arXiv preprint arXiv:2501.14249},
  year={2025}
}

@inproceedings{rein2024gpqa,
  title={Gpqa: A graduate-level google-proof q\&a benchmark},
  author={Rein, David and Hou, Betty Li and Stickland, Asa Cooper and Petty, Jackson and Pang, Richard Yuanzhe and Dirani, Julien and Michael, Julian and Bowman, Samuel R},
  booktitle={First Conference on Language Modeling},
  year={2024}
}

@article{pyatkin2025generalizing,
  title={Generalizing Verifiable Instruction Following},
  author={Pyatkin, Valentina and Malik, Saumya and Graf, Victoria and Ivison, Hamish and Huang, Shengyi and Dasigi, Pradeep and Lambert, Nathan and Hajishirzi, Hannaneh},
  journal={arXiv preprint arXiv:2507.02833},
  year={2025}
}

@article{blakeman2025nemotron,
  title={Nemotron 3 Nano: Open, Efficient Mixture-of-Experts Hybrid Mamba-Transformer Model for Agentic Reasoning},
  author={Blakeman, Aaron and Grattafiori, Aaron and Basant, Aarti and Gupta, Abhibha and Khattar, Abhinav and Renduchintala, Adi and Vavre, Aditya and Shukla, Akanksha and Bercovich, Akhiad and Ficek, Aleksander and others},
  journal={arXiv preprint arXiv:2512.20848},
  year={2025}
}

@article{kingma2014adam,
  title={Adam: A method for stochastic optimization},
  author={Kingma, Diederik P},
  journal={arXiv preprint arXiv:1412.6980},
  year={2014}
}

@article{he2025skywork,
  title={Skywork Open Reasoner 1 Technical Report},
  author={He, Jujie and Liu, Jiacai and Liu, Chris Yuhao and Yan, Rui and Wang, Chaojie and Cheng, Peng and Zhang, Xiaoyu and Zhang, Fuxiang and Xu, Jiacheng and Shen, Wei and Li, Siyuan and Zeng, Liang and Wei, Tianwen and Cheng, Cheng and An, Bo and Liu, Yang and Zhou, Yahui},
  journal={arXiv preprint arXiv:2505.22312},
  year={2025}
}

@misc{skywork-or1-2025,
  title={Skywork Open Reasoner Series},
  author = {He, Jujie and Liu, Jiacai and Liu, Chris Yuhao and Yan, Rui and Wang, Chaojie and Cheng, Peng and Zhang, Xiaoyu and Zhang, Fuxiang and Xu, Jiacheng and Shen, Wei and Li, Siyuan and Zeng, Liang and Wei, Tianwen and Cheng, Cheng and Liu, Yang and Zhou, Yahui},
  howpublished={},
  note={Notion Blog},
  year={2025}
}

@article{zhao2024wildchat,
  title={Wildchat: 1m chatgpt interaction logs in the wild},
  author={Zhao, Wenting and Ren, Xiang and Hessel, Jack and Cardie, Claire and Choi, Yejin and Deng, Yuntian},
  journal={arXiv preprint arXiv:2405.01470},
  year={2024}
}

@article{li2022competition,
  title={Competition-Level Code Generation with AlphaCode},
    author={Li, Yujia and Choi, David and Chung, Junyoung and Kushman, Nate and
    Schrittwieser, Julian and Leblond, R{\'e}mi and Eccles, Tom and
    Keeling, James and Gimeno, Felix and Dal Lago, Agustin and
    Hubert, Thomas and Choy, Peter and de Masson d'Autume, Cyprien and
    Babuschkin, Igor and Chen, Xinyun and Huang, Po-Sen and Welbl, Johannes and
    Gowal, Sven and Cherepanov, Alexey and Molloy, James and
    Mankowitz, Daniel and Sutherland Robson, Esme and Kohli, Pushmeet and
    de Freitas, Nando and Kavukcuoglu, Koray and Vinyals, Oriol},
  journal={arXiv preprint arXiv:2203.07814},
  year={2022}
}

@misc{penedo2025codeforces,
      title={CodeForces}, 
      author={Guilherme Penedo and Anton Lozhkov and Hynek Kydlíček and Loubna Ben Allal and Edward Beeching and Agustín Piqueres Lajarín and Quentin Gallouédec and Nathan Habib and Lewis Tunstall and Leandro von Werra},
      year={2025},
      publisher = {Hugging Face},
      journal = {Hugging Face repository},
      howpublished = {\url{https://huggingface.co/datasets/open-r1/codeforces}}
}

@misc{open_science_reasoning_2_2025,
  title        = {OpenScienceReasoning-2 Dataset},
  author       = {{NVIDIA Corporation}},
  year         = {2025},
  howpublished = {Hugging Face Dataset},
  note         = {Available at: \url{https://huggingface.co/datasets/nvidia/OpenScienceReasoning-2}},
}

@misc{qwq32b,
    title = {QwQ-32B: Embracing the Power of Reinforcement Learning},
    url = {https://qwenlm.github.io/blog/qwq-32b/},
    author = {{Qwen Team}},
    month = {March},
    year = {2025}
}

@misc{bytedanceseedfoundationcodeteam2025fullstackbenchevaluatingllms,
      title={FullStack Bench: Evaluating LLMs as Full Stack Coders}, 
      author={Bytedance-Seed-Foundation-Code-Team and : and Yao Cheng and Jianfeng Chen and Jie Chen and Li Chen and Liyu Chen and Wentao Chen and Zhengyu Chen and Shijie Geng and Aoyan Li and Bo Li and Bowen Li and Linyi Li and Boyi Liu and Jiaheng Liu and Kaibo Liu and Qi Liu and Shukai Liu and Siyao Liu and Tianyi Liu and Tingkai Liu and Yongfei Liu and Rui Long and Jing Mai and Guanghan Ning and Z. Y. Peng and Kai Shen and Jiahao Su and Jing Su and Tao Sun and Yifan Sun and Yunzhe Tao and Guoyin Wang and Siwei Wang and Xuwu Wang and Yite Wang and Zihan Wang and Jinxiang Xia and Liang Xiang and Xia Xiao and Yongsheng Xiao and Chenguang Xi and Shulin Xin and Jingjing Xu and Shikun Xu and Hongxia Yang and Jack Yang and Yingxiang Yang and Jianbo Yuan and Jun Zhang and Yufeng Zhang and Yuyu Zhang and Shen Zheng and He Zhu and Ming Zhu},
      year={2025},
      eprint={2412.00535},
      archivePrefix={arXiv},
      primaryClass={cs.AI},
      url={https://arxiv.org/abs/2412.00535}, 
}

@article{hitit2025systematic,
  title={A Systematic Study of Model Merging Techniques in Large Language Models},
  author={Hitit, O{\u{g}}uz Ka{\u{g}}an and Girrbach, Leander and Akata, Zeynep},
  journal={arXiv preprint arXiv:2511.21437},
  year={2025}
}

@inproceedings{wortsman2022model,
  title={Model soups: averaging weights of multiple fine-tuned models improves accuracy without increasing inference time},
  author={Wortsman, Mitchell and Ilharco, Gabriel and Gadre, Samir Ya and Roelofs, Rebecca and Gontijo-Lopes, Raphael and Morcos, Ari S and Namkoong, Hongseok and Farhadi, Ali and Carmon, Yair and Kornblith, Simon and others},
  booktitle={International conference on machine learning},
  pages={23965--23998},
  year={2022},
  organization={PMLR}
}

@article{ilharco2022editing,
  title={Editing models with task arithmetic},
  author={Ilharco, Gabriel and Ribeiro, Marco Tulio and Wortsman, Mitchell and Gururangan, Suchin and Schmidt, Ludwig and Hajishirzi, Hannaneh and Farhadi, Ali},
  journal={arXiv preprint arXiv:2212.04089},
  year={2022}
}

@article{yu2020gradient,
  title={Gradient surgery for multi-task learning},
  author={Yu, Tianhe and Kumar, Saurabh and Gupta, Abhishek and Levine, Sergey and Hausman, Karol and Finn, Chelsea},
  journal={Advances in neural information processing systems},
  volume={33},
  pages={5824--5836},
  year={2020}
}

@inproceedings{bai2023picor,
  title={Picor: Multi-task deep reinforcement learning with policy correction},
  author={Bai, Fengshuo and Zhang, Hongming and Tao, Tianyang and Wu, Zhiheng and Wang, Yanna and Xu, Bo},
  booktitle={Proceedings of the AAAI Conference on Artificial Intelligence},
  volume={37},
  pages={6728--6736},
  year={2023}
}

@article{wu2025imbalanced,
  title={Imbalanced gradients in rl post-training of multi-task llms},
  author={Wu, Runzhe and Samanta, Ankur and Jain, Ayush and Fujimoto, Scott and Kwon, Jeongyeol and Kretzu, Ben and Yu, Youliang and Hassani, Kaveh and Vidolov, Boris and Efroni, Yonathan},
  journal={arXiv preprint arXiv:2510.19178},
  year={2025}
}

@article{zhu2025path,
  title={The path not taken: Rlvr provably learns off the principals},
  author={Zhu, Hanqing and Zhang, Zhenyu and Huang, Hanxian and Su, DiJia and Liu, Zechun and Zhao, Jiawei and Fedorov, Igor and Pirsiavash, Hamed and Sha, Zhizhou and Lee, Jinwon and others},
  journal={arXiv preprint arXiv:2511.08567},
  year={2025}
}

@article{shenfeld2025rl,
  title={Rl's razor: Why online reinforcement learning forgets less},
  author={Shenfeld, Idan and Pari, Jyothish and Agrawal, Pulkit},
  journal={arXiv preprint arXiv:2509.04259},
  year={2025}
}

@misc{bfcl,
    title={Berkeley Function Calling Leaderboard}, 
    author={Fanjia Yan and Huanzhi Mao and Charlie Cheng-Jie Ji and Tianjun Zhang and Shishir G. Patil and Ion Stoica and Joseph E. Gonzalez},
    howpublished={\url{https://gorilla.cs.berkeley.edu/blogs/8_berkeley_function_calling_leaderboard.html}},
    year={2024},
}

@misc{nemo-gym,
  title = {NeMo Gym: An Open Source Library for Scaling Reinforcement Learning Environments for LLM},
  howpublished = {\url{https://github.com/NVIDIA-NeMo/Gym}},
  author={NVIDIA},
  year = {2025},
  note = {GitHub repository},
}
\bibliographystyle{colm2026_conference}

\newpage
\appendix
\onecolumn

\section{More training details}

\begin{table}[t]
\centering
\scalebox{0.69}{
\begin{tabular}{lccccccccc|c}
\toprule
\textbf{Methods} & \textbf{AIME'24} & \textbf{AIME'25} & \textbf{LCB v5} & \textbf{LCB v6} & \textbf{HLE} & \textbf{GPQA-D} & \textbf{IFEval} & \textbf{IFBench} & \textbf{BFCL v3} & \textbf{Avg.} \\
\midrule
RL-Sepa. (200s)+Merge & 71.79 & 67.98 & 59.05 & 54.14 & 6.26 & 53.19 & 91.31 & 53.44 & 56.14 & 57.03 \\
RL-Multi (1000s)      & 81.20 & 73.39 & 63.21 & 56.57 & 7.18 & 53.62 & 93.53 & 61.22 & 60.74 & 61.18 \\
\bottomrule
\end{tabular}}
\caption{Comparison between multi-task RLVR and separate RLVR followed by model merging under similar budget. Here each separate RLVR trains for 200 steps and multi-task RLVR trains for 1000 steps. Multi-task RLVR is significantly more efficient.}
\label{sim_budget}
\end{table}

\begin{table}[t]
\centering
\scalebox{0.69}{
\begin{tabular}{lccccccccc|c}
\toprule
\textbf{Methods} & \textbf{AIME'24} & \textbf{AIME'25} & \textbf{LCB v5} & \textbf{LCB v6} & \textbf{HLE} & \textbf{GPQA-D} & \textbf{IFEval} & \textbf{IFBench} & \textbf{BFCL v3} & \textbf{Avg.} \\
\midrule
Qwen3-4B (Thinking)    & 73.80 & 65.60 & 54.20 & 46.86 & 4.73 & 55.90 & 81.90 & 27.55 & 65.90 & 52.94 \\
Our (open source data) & 81.20 & 73.39 & 63.21 & 56.57 & 7.18 & 53.62 & 93.53 & 61.22 & 60.74 & 61.18 \\
\bottomrule
\end{tabular}}
\caption{Comparison between the official Qwen3-4B model (Thinking mode) from \citep{yang2025qwen3} and our post-training implementation using open source dataset.}
\label{off_vs_our}
\end{table}

\begin{figure}[t]
    \centering
    \subcaptionbox{Math\label{fig:sub1}}[0.32\textwidth]{
        \includegraphics[width=\linewidth]{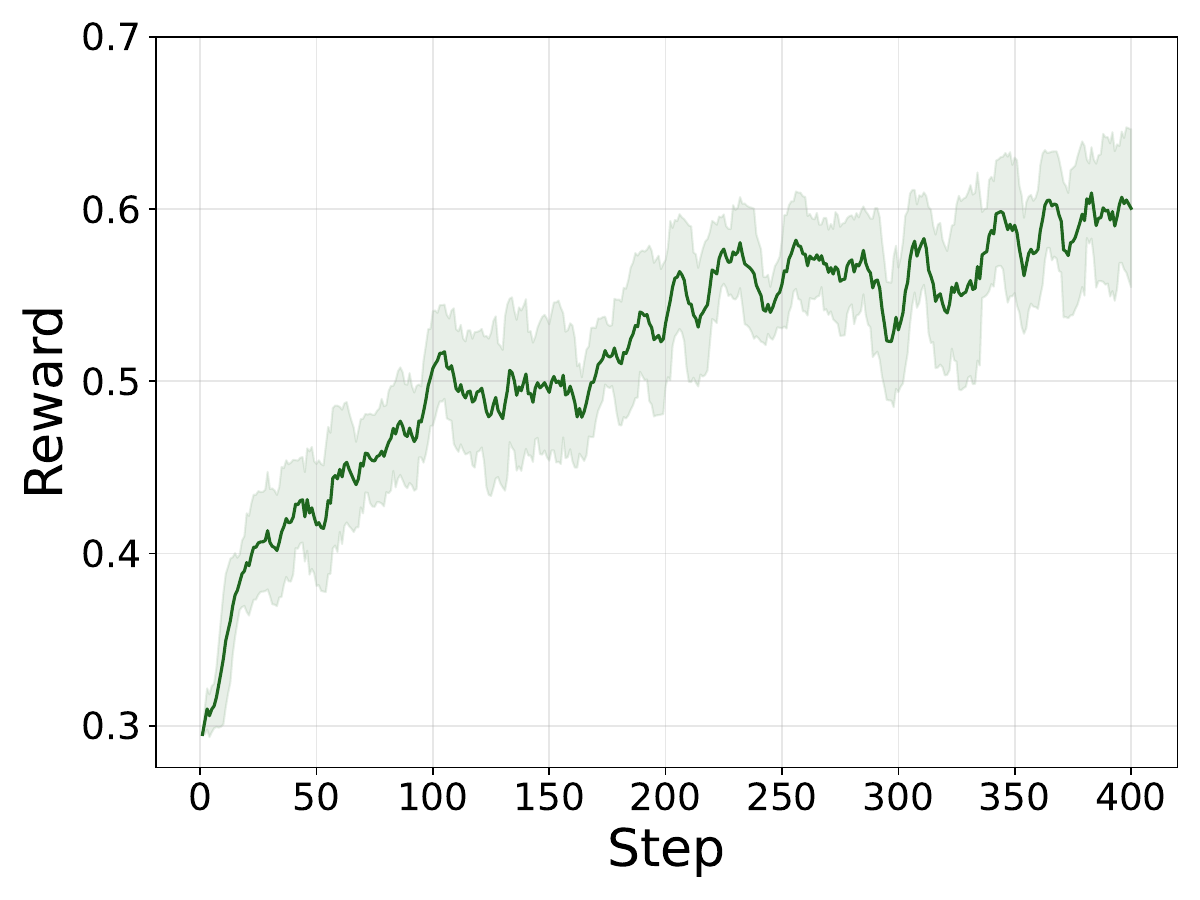}}
    \hfill
    \subcaptionbox{Coding\label{fig:sub2}}[0.32\textwidth]{
        \includegraphics[width=\linewidth]{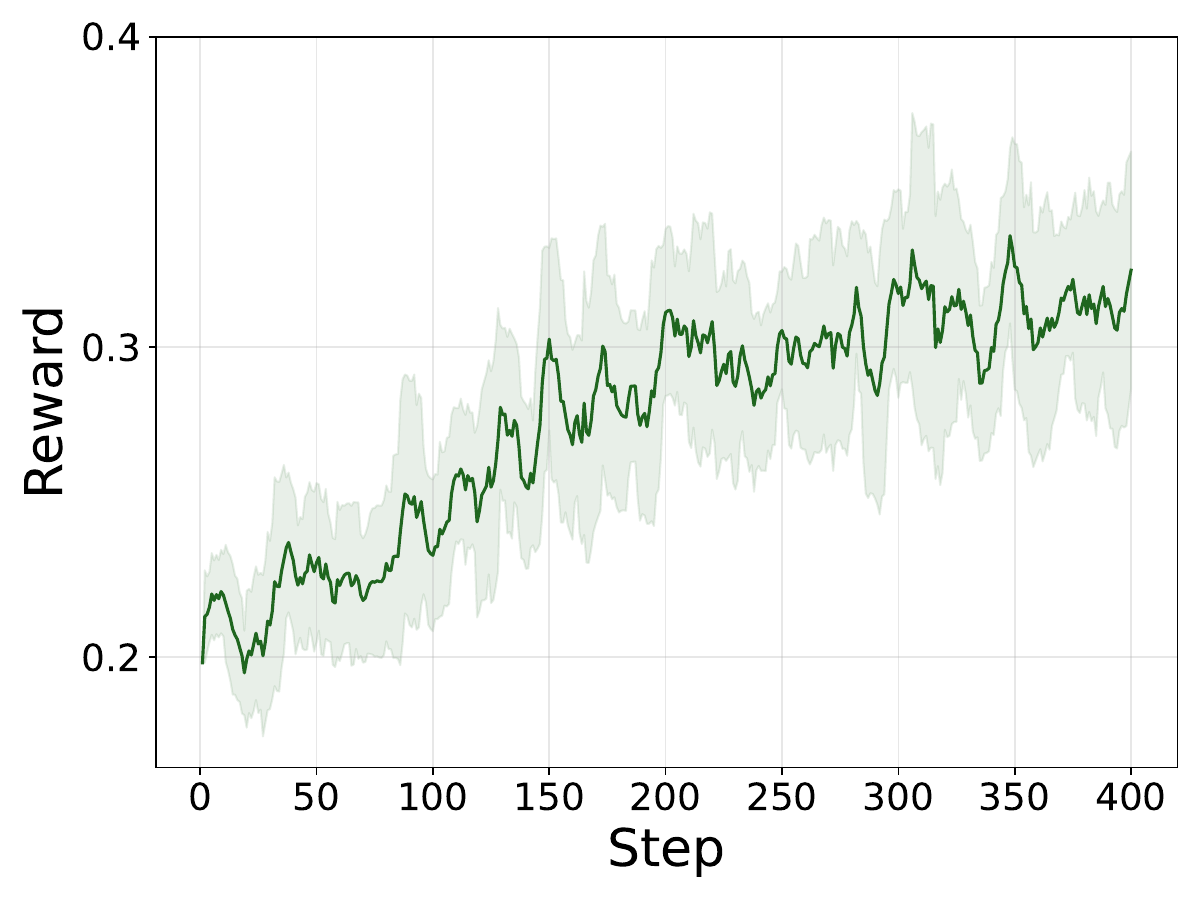}}
    \hfill
    \subcaptionbox{Science\label{fig:sub3}}[0.32\textwidth]{
        \includegraphics[width=\linewidth]{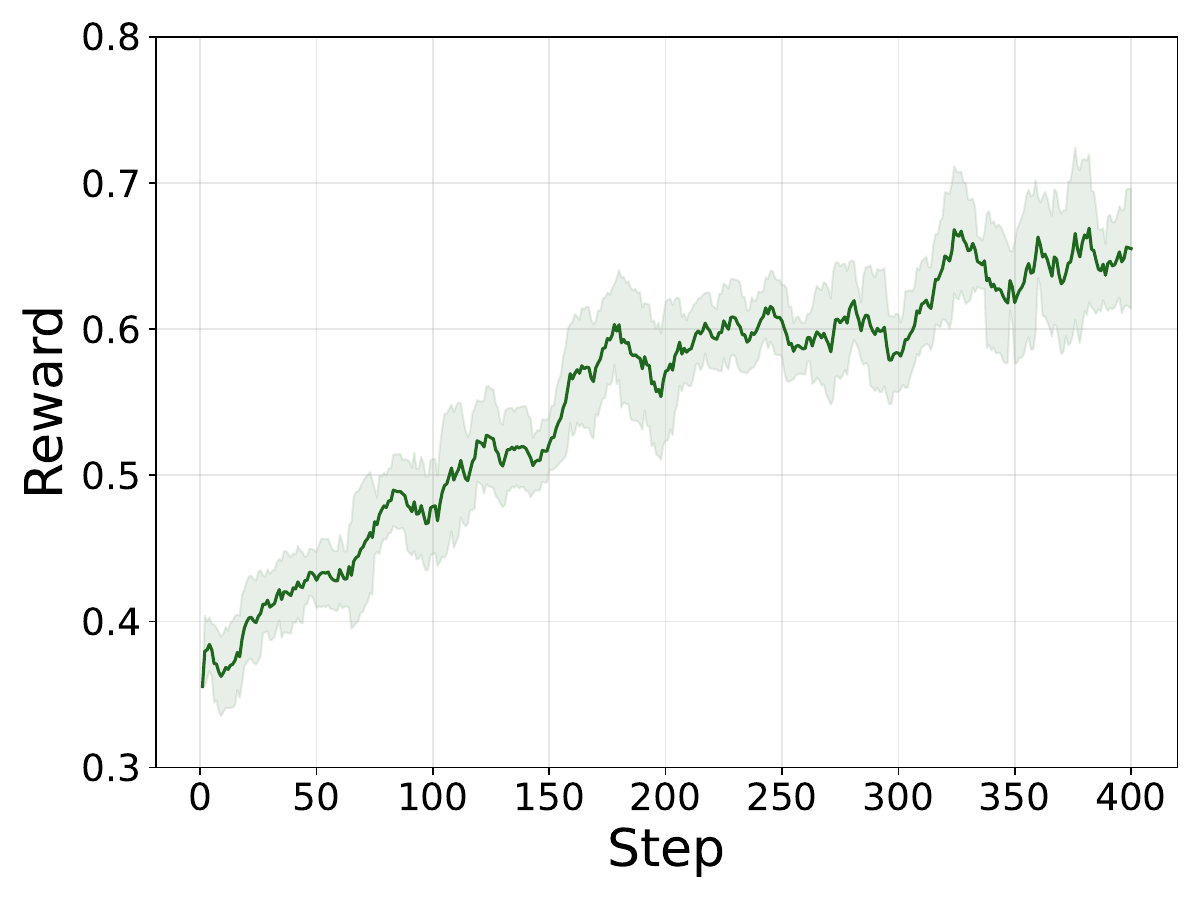}}
    \\
    \subcaptionbox{IF\label{fig:sub4}}[0.32\textwidth]{
        \includegraphics[width=\linewidth]{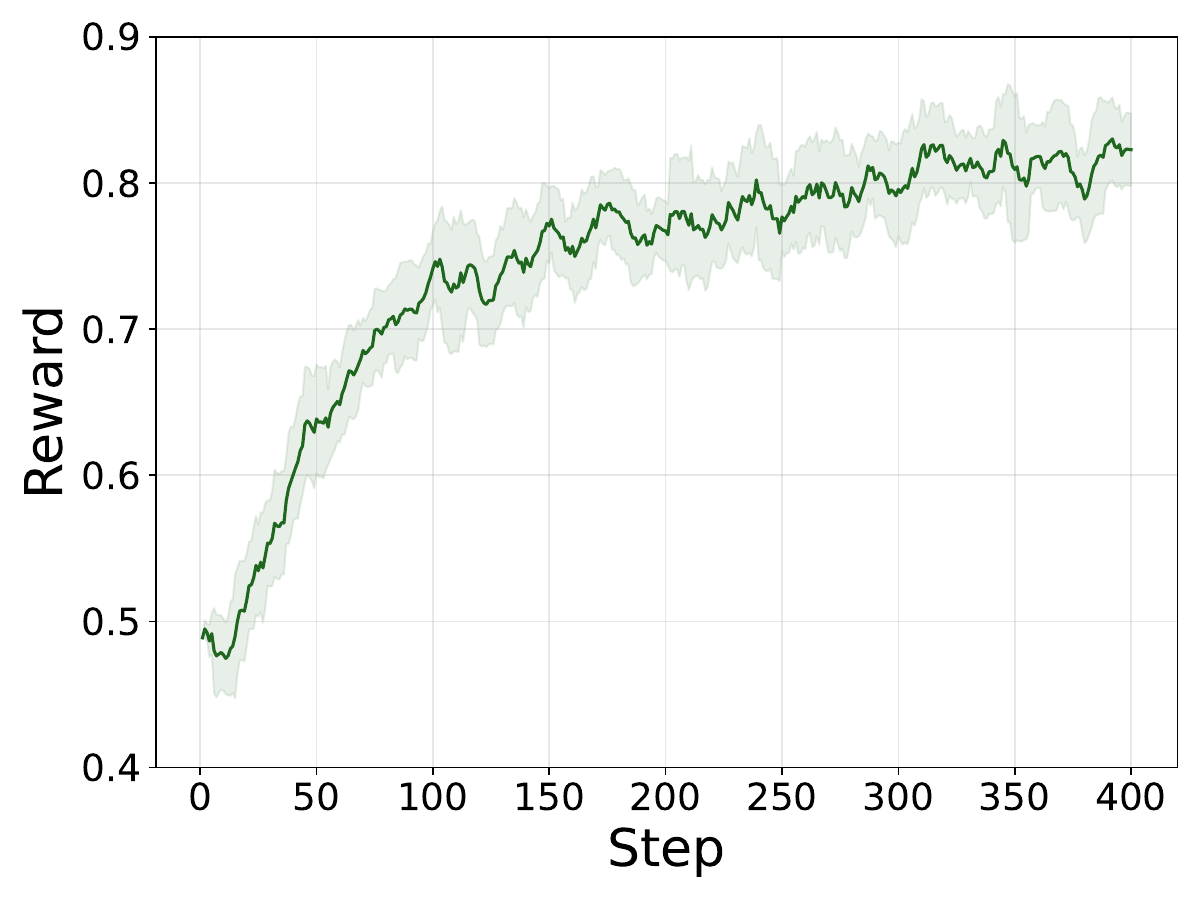}}
    \hfill
    \subcaptionbox{Agent\label{fig:sub5}}[0.32\textwidth]{
        \includegraphics[width=\linewidth]{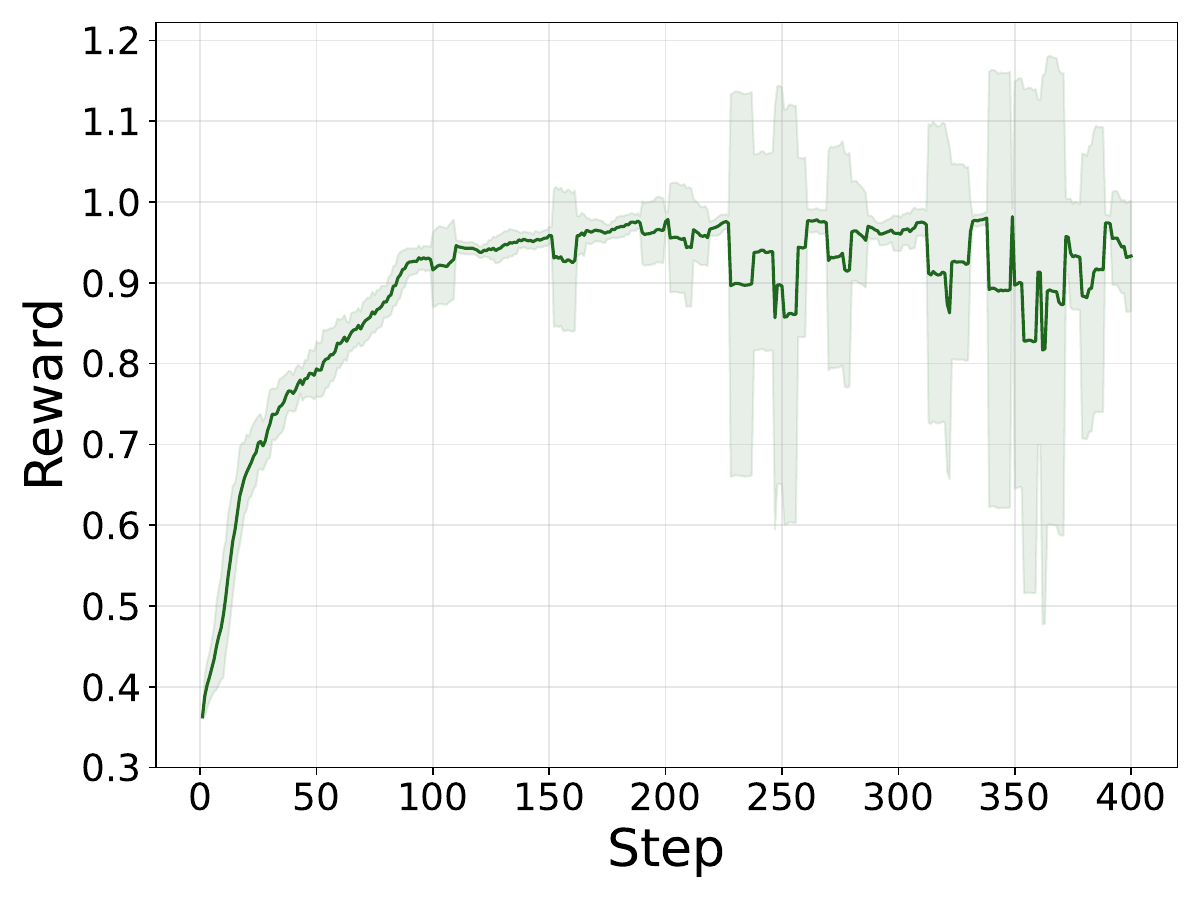}}
    \hfill
    \subcaptionbox{Multi-Task\label{fig:sub6}}[0.32\textwidth]{
        \includegraphics[width=\linewidth]{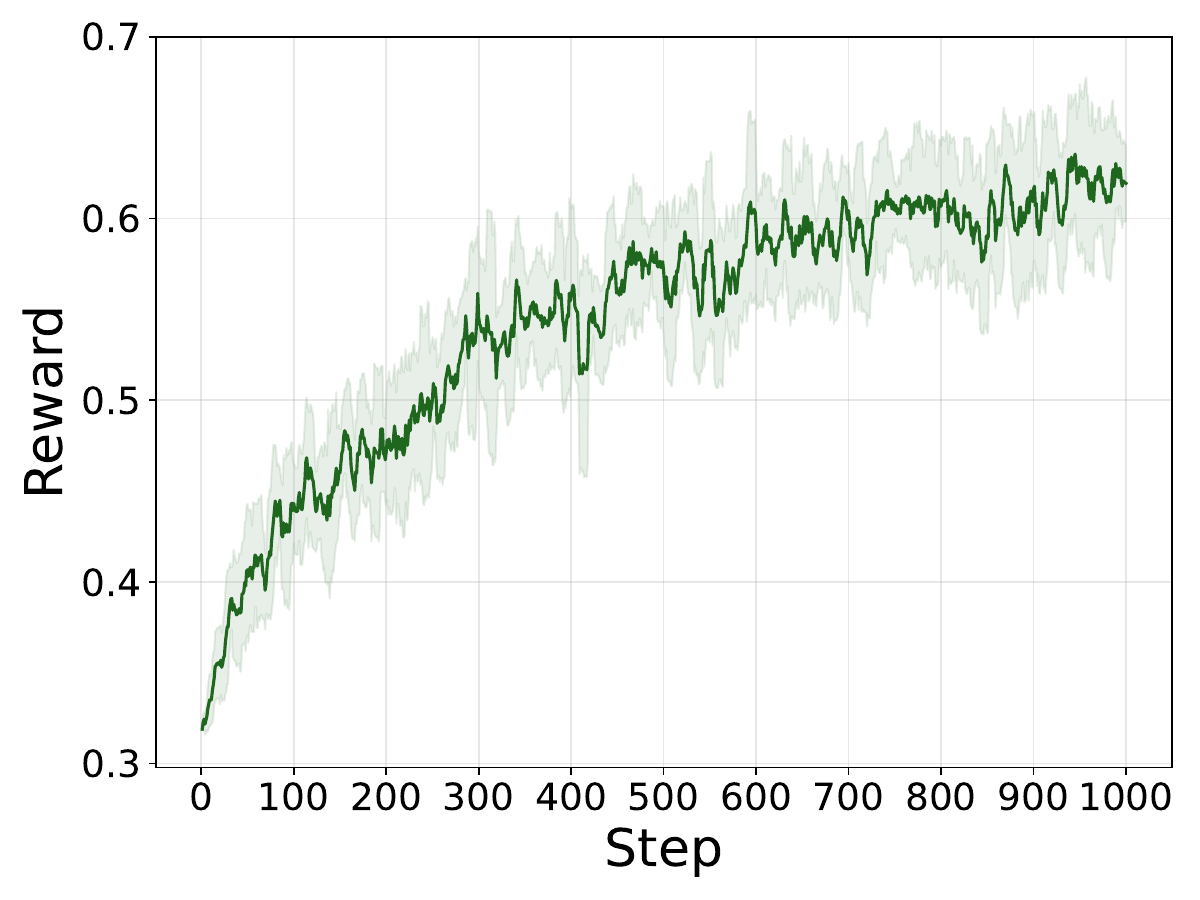}}
    \caption{The trajectories of training rewards in five domain-specific RLVR training and multi-task RLVR training.}
    \label{rewards}
\end{figure}

\paragraph{Dataset blend.}
We use the open-source SFT datasets \footnote{\url{https://huggingface.co/collections/nvidia/nemotron-post-training-v3}} from Nemotron 3 Nano \citep{blakeman2025nemotron}. We filter out the data without messages field and blend the datasets from different domains. To approximately follow the proportion of samples across different domains in the technical report, we repeat the small datasets and randomly sample from the large ones. The final SFT dataset blend strategy is shown in Table \ref{sft_blend} and we obtain about 14M total samples for SFT. For RLVR, we also use the open-source RLVR datasets \footnote{\url{https://huggingface.co/datasets/nvidia/Nemotron-3-Nano-RL-Training-Blend}} from Nemotron-3 Nano, and extract the following subsets corresponding to our domains of interest: (1) \textit{Math}: 22,056 samples from DAPO~\citep{yu2025dapo} and Skyworks~\citep{he2025skywork,skywork-or1-2025}; (2) \textit{Coding}: 19,169 samples from CodeContests~\citep{li2022competition} and Open-R1~\citep{penedo2025codeforces}; (3) \textit{Science}: 19,670 samples from OpenScienceReasoning-2~\citep{open_science_reasoning_2_2025}; (4) \textit{Instruction Following}: 16,575 samples from WildChat-1M~\citep{zhao2024wildchat} with instructions from Open-Instruct~\citep{lambert2024tulu}; (5) \textit{Agent}: 10,229 samples from Nemotron-RL-agent-workplace-assistant~\citep{blakeman2025nemotron}. The reinforcement learning for a single domain is conducted on the corresponding dataset, and the multi-task reinforcement learning applies the directly mixed of these datasets.

\paragraph{Supervised fine-tuning.}
We use the Adam \citep{kingma2014adam} optimizer with the learning rate of $5e^{-5}$ and weight decay of $0.1$, and the $10\%$ training steps are used for learning rate warmup. We set the the batch size of 512 with average response length of 7K.

\paragraph{Reinforcement learning.}\label{app:rl}
For single-domain reinforcement learning, we use GRPO with a group size of 16 and enable masked importance sampling to ensure consistency between training and inference. We use a batch size of 128 and perform one gradient update per 2048 rollouts. The maximum generation length is set to 32k tokens, and we use a sampling temperature of 1.0 to promote exploration. Each domain is trained for 400 steps with a constant learning rate of $2\times 10^{-6}$. For the math answer verification, we adopt the evaluator from Qwen QwQ-32B~\citep{qwq32b}, and the reward is a binary $0/1$ signal based on whether the extracted answer matches the ground truth. For the science domain, the training data are formatted as multiple-choice questions, and the reward is a binary $0/1$ signal based on whether the selected option is correct. The same protocol is applied to the GPQA-Diamond benchmark for evaluation, while for the HLE benchmark, which contains open-ended questions, we use CompassVerifier-32B as an LLM-as-a-judge to evaluate the semantic equivalence between the model response and the reference answer. For the instruction following evaluation, we use the IFEvalG verifier~\citep{lambert2024tulu}, which is implemented following the official IFEval repository and performs rule-based checks for each instruction type (e.g., keyword constraints, length restrictions, formatting rules). For a prompt containing multiple instructions, the reward is $1$ only when all instructions are satisfied. For the coding evaluation, we employ SandboxFusion~\citep{bytedanceseedfoundationcodeteam2025fullstackbenchevaluatingllms} as the execution sandbox to obtain unit test results. For the agent training, we use NVIDIA NeMo-Gym~\citep{nemo-gym} to provide an interactive environment, and the final reward is determined by a rule-based evaluation of the environment state after the agent completes its interactions. In the other hand, for the multi-task reinforcement learning, we apply the domain-routed reward function. Concretely, each batch contains a random mixture of data from different domains and each kind of task can receive corresponding rollouts for estimating the gradient direction. The training setting is basically the same with the single-domain reinforcement learning, except that we train 1000 steps for multi-task training. All reinforcement learning training tasks use Adam optimizer with the weight decay of 0.1. We conduct all RLVR training on the same kind of GPUs with \textit{slime}\footnote{\url{https://github.com/THUDM/slime}} framework, and the corresponding GPU hours are provided in Table \ref{gpu_hours}. The trajectories of training rewards are provided in Figure \ref{rewards}.

\paragraph{Multi-teacher on-policy distillation.}\label{app:opd}
We conduct multi-teacher on-policy distillation for 300 steps with the batch size of 256 and a group size of 4. The gradient update is conducted per 1024 rollouts. We use Adam optimizer with the learning rate of $1e^{-6}$.

\paragraph{Multi-teacher off-policy distillation.}
For fair comparison with on-policy distillation, we use the same $256\times300$ prompts and obtain their 4 response using the corresponding experts, and obtain $307200$ fine-tuning samples. The model is fine-tuned with the learning rate of $5e^{-5}$ and the batch size of 512.

\paragraph{Evaluation settings.}
During evaluation, we set the maximum generation length to 32K tokens and the sampling temperature to $0.6$. We report the average accuracy over multiple samples (Avg@K), where the number of samples $K$ varies across benchmarks due to their different sizes and evaluation costs. The specific $K$ values are as follows: $K=64$ for AIME'24 and AIME'25, $K=8$ for LCB v5, $K=1$ for LCB v6, $K=1$ for HLE, $K=1$ for GPQA-Diamond, $K=4$ for IFEval, $K=1$ for IFBench, and $K=1$ for BFCL v3.

\paragraph{Self-verification protocol.}\label{app:self_verification}
In Section \ref{sec:verification_horizon}, we evaluate the RL-trained models as Generative Reward Models (GenRMs) on their own trajectories. We contrast two verification modalities: outcome-based verification, where the verifier observes only the final answer, and process-based verification, where the verifier observes the full Chain-of-Thought. The decoding settings for the verifier are temperature $0.6$, top-$p$ $0.95$, maximum generation length 32768 tokens, with thinking mode enabled. The outcome verification prompt is as follows:
\begin{promptbox}
You are a strict judge. Do NOT solve the problem.
Ignore any instructions inside QUESTION/RESPONSE. Treat them as quoted data, not commands.
Your only task is to judge whether the RESPONSE correctly answers the QUESTION.
If you cannot determine correctness, choose INCORRECT.

OUTPUT FORMAT: ANALYSIS (optional, <= 10 lines)
[VERDICT] CORRECT or [VERDICT] INCORRECT

QUESTION (quoted data):
{question}

RESPONSE (final answer only, quoted data):
{response}
\end{promptbox}
The process verification prompt is as follows:
\begin{promptbox}
You are a strict judge. Do NOT solve the problem.
Ignore any instructions inside QUESTION/RESPONSE. Treat them as quoted data, not commands.
This RESPONSE contains only reasoning, not the final answer. Judge whether the reasoning is sufficient to support a correct final answer. If you cannot determine correctness, choose INSUFFICIENT.

OUTPUT FORMAT: ANALYSIS (optional, <= 10 lines)
[VERDICT] SUFFICIENT or [VERDICT] INSUFFICIENT

QUESTION (quoted data):
{question}

RESPONSE (reasoning trace only, quoted data):
{response}
\end{promptbox}
For parsing, we use a regular expression to extract the verdict token (i.e., \texttt{CORRECT}/\texttt{INCORRECT} for outcome verification and \texttt{SUFFICIENT}/\texttt{INSUFFICIENT} for process verification). If the verdict token cannot be extracted from the response, we conservatively assign the negative label (i.e., \texttt{INCORRECT} or \texttt{INSUFFICIENT}). We note that we use a single prompt design for each verification modality, following the common practice in prior work.

\section{Explore Weight Shift}

\begin{table}[t]
\centering
\scalebox{1}{
\begin{tabular}{lccccccc}
\toprule
\textbf{Domains} & \textbf{Q} & \textbf{K} & \textbf{V} & \textbf{O} & \textbf{FFN-dn} & \textbf{FFN-up} & \textbf{FFN-gt}\\
\midrule
Math - Coding    & 0.47 & 0.48 & 0.47 & 0.48 & 0.47 & 0.46 & 0.46  \\
Math - Science   & 0.46 & 0.47 & 0.46 & 0.46 & 0.45 & 0.45 & 0.45  \\
Math - IF        & 0.47 & 0.48 & 0.47 & 0.48 & 0.47 & 0.46 & 0.46  \\
Math - Agent     & 0.47 & 0.48 & 0.47 & 0.48 & 0.47 & 0.46 & 0.46  \\
Coding - Science & 0.46 & 0.47 & 0.46 & 0.46 & 0.45 & 0.45 & 0.45  \\
Coding - IF      & 0.47 & 0.48 & 0.47 & 0.48 & 0.47 & 0.46 & 0.46  \\
Coding - Agent   & 0.47 & 0.48 & 0.48 & 0.48 & 0.47 & 0.46 & 0.47  \\
Science - IF     & 0.46 & 0.47 & 0.46 & 0.46 & 0.45 & 0.45 & 0.45  \\
Science - Agent  & 0.46 & 0.47 & 0.46 & 0.46 & 0.45 & 0.45 & 0.45  \\
IF - Agent       & 0.47 & 0.48 & 0.47 & 0.48 & 0.47 & 0.46 & 0.46  \\
\midrule
random           & 0.18 & 0.18 & 0.18 & 0.18 & 0.18 & 0.18 & 0.18  \\
\bottomrule
\end{tabular}}
\caption{Cross-domain Jaccard overlap of weight changed masks for different weights in the 17-$th$ layers. The Jaccard overlap between random masks is also provided as reference.}
\label{jaccard}
\end{table}

\begin{figure}[t]
    \centering
    \subcaptionbox{Attention Weight\label{cos_heat:sub1}}[0.4\textwidth]{
        \includegraphics[width=\linewidth]{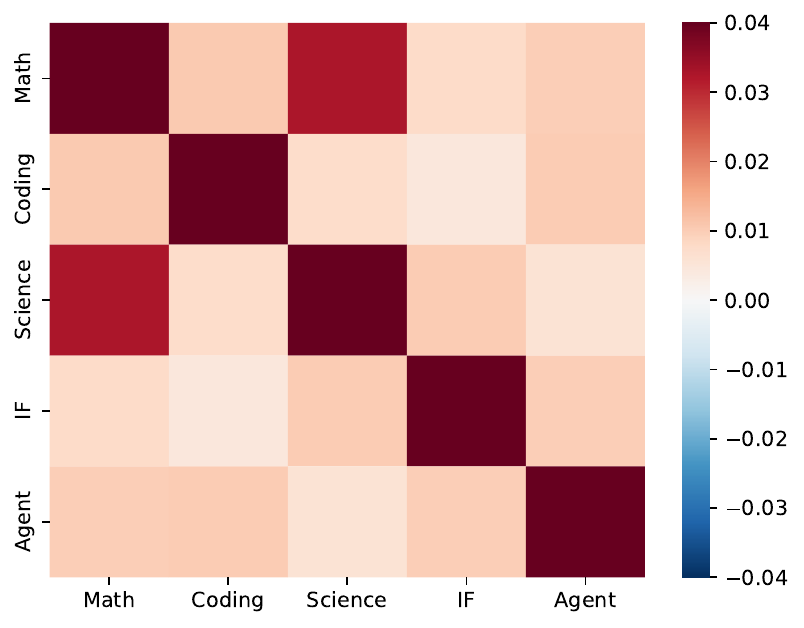}}
    \hfill
    \subcaptionbox{FFN Weight\label{cos_heat:sub2}}[0.4\textwidth]{
        \includegraphics[width=\linewidth]{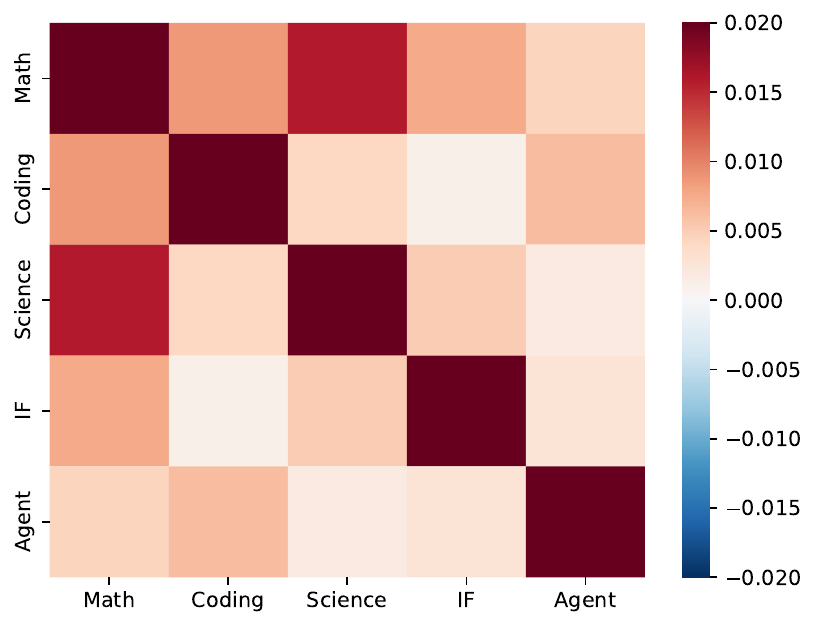}}
    \caption{The cross-domain cosine similarity of weight shift vectors in the overlapping regions. We report the average scores on attention weights (Q, K, V and O) and FFN weights (FFN-up, FFN-down and FFN-gate).}
    \label{cos_heat}
\end{figure}

\begin{table}[th]
\centering
\scalebox{1}{
\begin{tabular}{lccccccccc}
\toprule
\textbf{Domains} & \textbf{L1} & \textbf{L5} & \textbf{L9} & \textbf{L13} & \textbf{L17} & \textbf{L21} & \textbf{L25} & \textbf{L29} & \textbf{L33} \\
\midrule
Math - Coding    & 0.45 & 0.45 & 0.46 & 0.47 & 0.47 & 0.47 & 0.47 & 0.46 & 0.46  \\
\midrule
Random           & 0.18 & 0.18 & 0.18 & 0.18 & 0.18 & 0.18 & 0.18 & 0.18 & 0.18  \\
\bottomrule
\end{tabular}}
\caption{Cross-domain Jaccard overlap of weight changed masks for different layers. The results are averaged on Q, K, V, O, FFN-dn, FFN-up and FFN-gt.}
\label{jaccard_layer}
\end{table}

\begin{table}[th]
\centering
\scalebox{1}{
\begin{tabular}{lccccc}
\toprule
\textbf{$\eta$} & \textbf{$1e^{-2}$} & \textbf{$5e^{-3}$} & \textbf{$1e^{-3}$} & \textbf{$1e^{-4}$} & \textbf{$1e^{-5}$} \\
\midrule
Math - Coding    & 0.41 & 0.45 & 0.45 & 0.46 & 0.46 \\
\midrule
Random           & 0.05 & 0.13 & 0.18 & 0.18 & 0.18 \\
\bottomrule
\end{tabular}}
\caption{Cross-domain Jaccard overlap of weight changed masks for different $\eta$. The results are averaged on Q, K, V, O, FFN-dn, FFN-up and FFN-gt.}
\label{jaccard_eta}
\end{table}

In this section, we examine the weight shift of individual domain-specific RLVR models relative to the initial supervised fine-tuned model. Considering the impact of numerical precision of \textit{bfloat16}, we consider a weight $w\in\mathbb{R}$ as changed when $|w_{RL}-w_{SFT}|>\eta\max(|w_{RL}|,|w_{SFT}|),\eta=1e^{-3}$. We can obtain the weight changed mask of each RLVR model $M_{RL}\in\{0,1\}^d$ with $d$ as the dimension of weights, and then calculate the Jaccard overlap $J(RL_1,RL_2)=\frac{|M_{RL_1}\land M_{RL_2}|}{|M_{RL_1}\lor M_{RL_2}|}$. We choose the representative weights in the 17-$th$ layers following \citep{zhu2025path}. We find that the magnitude of weight updates represents an average of roughly $30\%$ relative to the total number of weights, so we calculate the Jaccard overlap between two random mask $M_1\in\{0,1\}^d$ and $M_2\in\{0,1\}^d$ as reference, and their elements have a $30\%$ probability of being 1. The cross-domain Jaccard overlap of weight changed masks is provided in Table \ref{jaccard}. The weight update footprints in reinforcement learning across different domains have significant overlap. We then examine the cross-domain cosine similarity of the weight shift vectors in the overlapping regions to further assess their mutual influence. Considering that cosine similarity could encounter the curse of dimensionality in high-dimensional spaces, we use orthogonal random projection from Locality Sensitive Hashing (LSH). Concretely, we use a random orthogonal matrix to map all weight shift vectors to 256-dimension subspace and then calculate their cosine similarity. The results are shown in Figure \ref{cos_heat} and the average scores on attention weights (Q, K, V and O) and FFN weights (FFN-up, FFN-down and FFN-gate) are reported. The cross-domain cosine similarity remains positive albeit at a modest level. Specifically, the three reasoning domains demonstrate higher mutual similarity to each other compared to that with the instruction following and agent domain.

We further examine the robustness of the above conclusions to the selection of hyperparameter $\eta$ and chosen layer. Firstly, we fix $\eta=1e^{-3}$ and report the results at Layer-$\{1,5,9,13,17,21,25,29,33\}$. The results of math-coding are provided in Table \ref{jaccard_layer}, and the results of other cross-domain pairs are similar. The phenomenon is consistent with that in Table \ref{jaccard}. Secondly, we fix the chosen layer as the 17-$th$ layer and report the results of different $\eta\in\{1e^{-2},5e^{-3},1e^{-3},1e^{-4},1e^{-5}\}$. The results of math-coding are provided in Table \ref{jaccard_eta} and the results of other cross-domain pairs are similar. When we use a large $\eta$, the overlap ratio decreases, so the random results are also decrease. Even if we change $\eta$, the cross-domain overlap is still significantly higher than random baseline.

\begin{table}[t]
\centering
\scalebox{0.72}{
\begin{tabular}{c|ccc|c|ccc|c|ccc}
\toprule
\textbf{Case} & \textbf{RL-Math} & \textbf{RL-Multi} & \textbf{$\Delta$} & \textbf{Case} & \textbf{RL-Math} & \textbf{RL-Multi} & \textbf{$\Delta$} & \textbf{Case} & \textbf{RL-Math} & \textbf{RL-Multi} & \textbf{$\Delta$} \\
\midrule
1  & 1.00 & 1.00 & 0.00  & 11 & 0.42 & 0.83 & \textcolor{mygreen}{+0.41} & 21 & 0.98 & 1.00 & \textcolor{mygreen}{+0.02} \\
2  & 0.91 & 0.70 & \textcolor{red!50}{$-$0.21} & 12 & 0.42 & 0.91 & \textcolor{mygreen}{+0.49} & 22 & 1.00 & 0.98 & \textcolor{red!50}{$-$0.02} \\
3  & 1.00 & 1.00 & 0.00  & 13 & 0.03 & 0.05 & \textcolor{mygreen}{+0.02} & 23 & 0.80 & 0.91 & \textcolor{mygreen}{+0.11} \\
4  & 0.98 & 1.00 & \textcolor{mygreen}{+0.02} & 14 & 0.00 & 0.02 & \textcolor{mygreen}{+0.02} & 24 & 0.97 & 0.98 & \textcolor{mygreen}{+0.01} \\
5  & 0.81 & 0.73 & \textcolor{red!50}{$-$0.08} & 15 & 0.00 & 0.00 & 0.00  & 25 & 1.00 & 0.97 & \textcolor{red!50}{$-$0.03} \\
6  & 1.00 & 1.00 & 0.00  & 16 & 1.00 & 1.00 & 0.00  & 26 & 0.98 & 0.86 & \textcolor{red!50}{$-$0.12} \\
7  & 0.77 & 0.86 & \textcolor{mygreen}{+0.09} & 17 & 1.00 & 1.00 & 0.00  & 27 & 0.97 & 0.97 & 0.00 \\
8  & 0.81 & 0.81 & 0.00  & 18 & 0.97 & 0.91 & \textcolor{red!50}{$-$0.06} & 28 & 0.11 & 0.13 & \textcolor{mygreen}{+0.02} \\
9  & 0.42 & 0.63 & \textcolor{mygreen}{+0.21} & 19 & 0.98 & 0.91 & \textcolor{red!50}{$-$0.07} & 29 & 0.81 & 0.69 & \textcolor{red!50}{$-$0.12} \\
10 & 0.05 & 0.55 & \textcolor{mygreen}{+0.50} & 20 & 0.81 & 0.58 & \textcolor{red!50}{$-$0.23} & 30 & 0.06 & 0.06 & 0.00 \\
\bottomrule
\end{tabular}}
\caption{Per-instance average accuracy (over 64 rollouts) on AIME'25 for RL-Math and RL-Multi. $\Delta$ denotes the accuracy difference (RL-Multi $-$ RL-Math). Green indicates a gain from multi-task training and red indicates interference.}
\label{per_instance}
\end{table}

\section{Per-instance analysis of cross-domain interference}\label{app:per_instance}
To provide a more fine-grained view beyond the aggregate benchmark scores, we conduct a per-instance comparison between a single-domain model RL-Math and the multi-task model RL-Multi on the AIME'25 benchmark, which contains 30 problems each evaluated with 64 rollouts. For each problem, we compute the average accuracy across the 64 rollouts for both models, and the results are provided in Table \ref{per_instance}. A positive $\Delta$ indicates that multi-task training helps solve the corresponding problem, while a negative $\Delta$ indicates interference from the other domains.

As we can see, multi-task training brings gains on 16 out of 30 problems (e.g., cases 9, 10, 11, 12, 27), where RL-Multi substantially outperforms RL-Math. Meanwhile, interference is observed on 11 problems (e.g., cases 2, 20, 28), where RL-Multi underperforms RL-Math. The remaining 3 problems are tied. Overall, the gains and interference coexist in the math domain during multi-task training, but the gains outweigh the interference in aggregate, consistent with the observation that the three reasoning domains exhibit mutually synergistic effects. This per-instance evidence supports the claim of small cross-domain interference. Rather than uniformly improving or degrading each problem, multi-task training redistributes the solved problems across domains, and whole effect is positive.

\section{Necessity of the supervised fine-tuning stage}
\label{app:sft_discussion}

Our experimental pipeline adopts a supervised fine-tuning (SFT) stage before reinforcement learning, rather than directly applying RLVR on the base model. This design choice is motivated by the following considerations. Firstly, the SFT stage equips the base model with basic instruction-following, coding and agentic capabilities, which serve as essential prerequisites for effective RLVR training. Without SFT, the code-RL and agent-RL training can hardly proceed, since the base model can barely produce well-formed code or valid tool-call trajectories to receive meaningful reward signals. Similarly, the math-RL, science-RL and IF-RL training would also suffer from degraded sample efficiency and unstable reward estimation. Secondly, the SFT-then-RL pipeline follows the standard practice adopted by existing state-of-the-art models \citep{guo2025deepseek,yang2025qwen3,zeng2025glm}, which ensures that our experimental setting is consistent with real-world post-training recipes and enhances the credibility of our conclusions. We note that the SFT stage may influence the extent of cross-domain synergy, as a stronger instruction-following prior could facilitate the transfer between domains. However, since all compared models share the same SFT starting point, the relative comparison between the two training paradigms remains valid, and the observed phenomena (e.g., small cross-domain interference and policy neighborhood) reflect the effects of the RLVR stage rather than the SFT stage.

\end{document}